\pdfoutput=1
\documentclass[11pt]{article}

\usepackage[final]{acl}
\usepackage{times}
\usepackage{latexsym}

\usepackage[T1]{fontenc}
\usepackage[utf8]{inputenc}

\usepackage{microtype}

\usepackage{inconsolata}

\usepackage{graphicx}

\usepackage{hyperref}
\usepackage{url}
\usepackage{multirow}
\usepackage{booktabs}
\usepackage{adjustbox}
\usepackage{CJKutf8}
\usepackage{listings}
\usepackage{longtable}
\usepackage{array}
\usepackage{subcaption}
\usepackage{float}
\usepackage{tabularx}
\CJK{UTF8}{gbsn}
\usepackage{amsmath}
\usepackage{amsfonts}
\title{\textsc{VideoNorms}: Benchmarking Socio-Cultural Norm Understanding of
Video Language Models}

\author{Nikhil Reddy Varimalla$^{\circ}$\textsuperscript{\S}\textsuperscript{*}\quad
Yunfei Xu$^{\circ}$\thanks{Equal contribution.}\quad
Meng Fan Wang$^{\circ}$\quad
Arkadiy Saakyan$^{\circ}$\quad
Smaranda Muresan$^{\circ}$$^{\bullet}$\quad
\\ 
$^{\circ}$Department of Computer Science, Columbia University \hspace{8pt} \\
$^{\bullet}$Barnard College \\
\texttt{nv2415@columbia.edu}
}

\begin{document}
\maketitle

\begin{abstract}
As Video Large Language Models (VideoLLMs) are deployed globally, it is important to assess their ability to reason across cultural contexts. 
To advance cultural norm awareness evaluation in VideoLLMs, we introduce \textsc{VideoNorms}, a dataset of cultural norm annotations from popular US and Chinese TV shows annotated with adherence or violation labels and (non-)verbal evidence. 
Through a human-AI collaboration framework, each item was first annotated by a large VideoLLM, and then reviewed by at least three trained monocultural annotators with significant lived experience in the target culture, resulting in a dataset of over 3,000 human judgments. Human verification showed disparity in US and Chinese norm extraction performance, cautioning against fully automatic approaches cultures under-represented in training data.  
Hierarchical linear modeling analysis of $7$ open-weight VideoLLMs' performance revealed that: 1) models perform worse in Chinese compared to US, particularly for norm adherence prediction; 2) models have more difficulty in providing non-verbal evidence compared to verbal evidence for norm adherence/violation predictions. Ablation studies confirm video modality is indeed necessary for accurate performance, and scaling model size does not yield classification score improvements.
Our findings and data contribute to culturally grounded video model training and evaluation.
% \footnote{Code and data: \texttt{\href{https://github.com/nikhilreddy3/VideoNorms}{github.com/nikhilreddy3/VideoNorms}}}

\end{abstract}

\section{Introduction}

\begin{figure*}[ht!]
% \vspace{-2mm}
    \centering
    \includegraphics[width=0.9\textwidth]{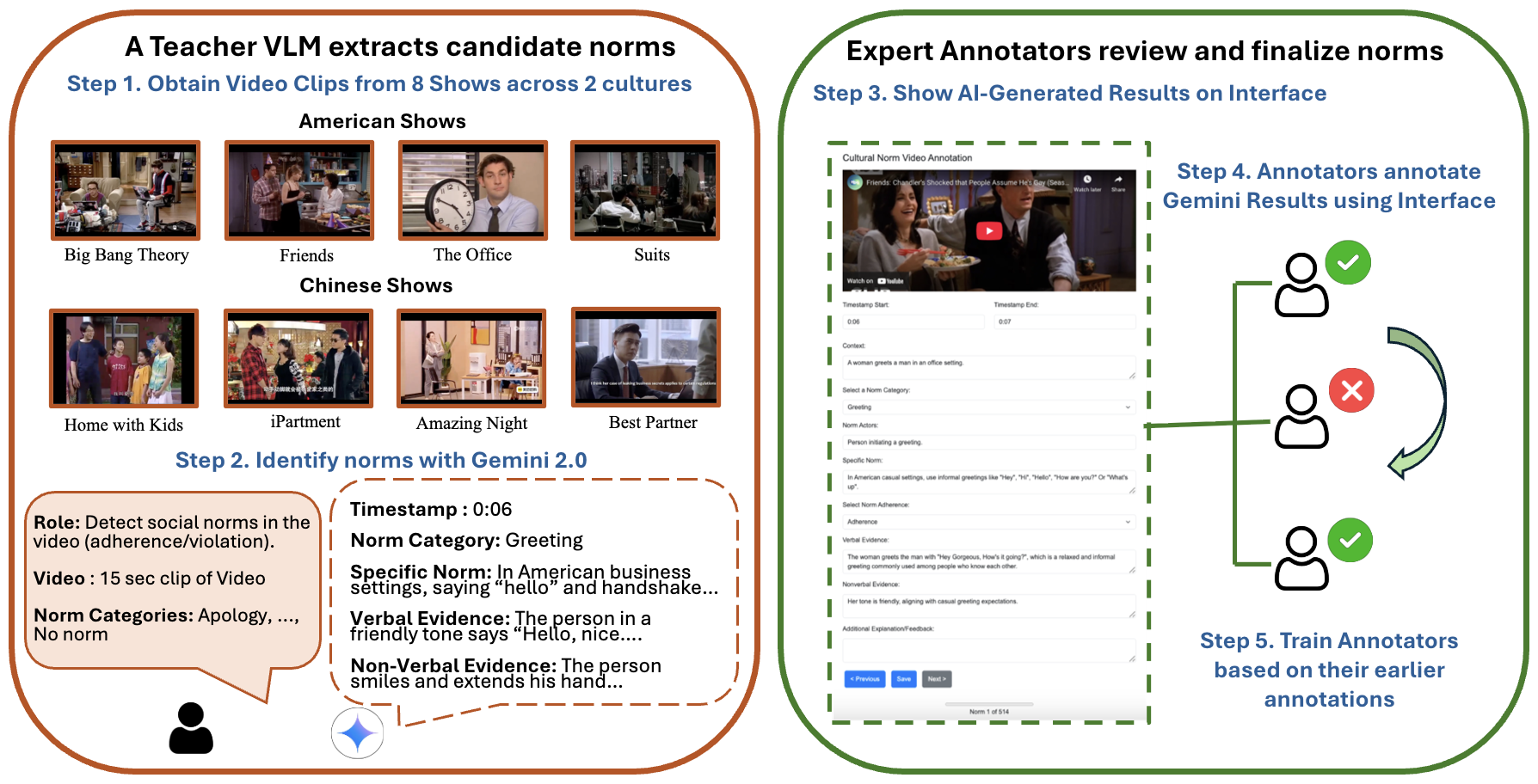}
    \caption{\textsc{VideoNorms} Dataset Construction: left panel shows teacher VideoLLM generations using prompting based on speech act and prior norm understanding; right panel shows the expert annotator editing process.}
    \label{fig:pipeline}
% \vspace{-2mm}    
\end{figure*}
Artificial Intelligence systems based on large language models (LLMs), including
video language models (VideoLLMs), are deployed globally, requiring adaptation to differing cultural contexts. However, cultural competence of VideoLLMs has not been given the same attention as general object/action recognition \citep{xu2016msrvtt}, temporal reasoning \citep{li2024mvbench,fu2025videomme, xiao2021nextqa, jang2017tgifqa}, or narrative understanding \citep{ tapaswi2016movieqa}. In the textual modality, frameworks for evaluating cultural competence are largely based on understanding social norms -- rules of thumb for human behavior, such as ``It is rude to run
a blender at 5am'' \citep{forbes2020socialchem, emelin2021moralstories, hendrycks2021ethics, ziems2023normbank} in differing cultural contexts \citep{shi-etal-2024-culturebank, huang-yang-2023-culturally, li-etal-2023-normdial, ch-wang-etal-2023-sociocultural}. In images, benchmarks testing the models' cultural knowledge \citep{winata-etal-2025-worldcuisines} and culturally-aligned image generation have been proposed \citep{nayak2025culturalframesassessingculturalexpectation}. In videos, recent work has %also 
explored norms grounded in the physical world \citep{rezaei-etal-2025-egonormia}. 

Expanding on these approaches, we propose to evaluate the cultural competence of VideoLLMs through their understanding of \emph{socio-cultural norms}: societal rules or standards that ``delineate an accepted and appropriate behavior within a culture'' \citep{apa_cultural_norm}, when the input modality is a video. For example, does the video language model understand that when two people introduce themselves and shake hands, they adhere to the greeting norms in US culture? This is particularly challenging in the multimodal context, where understanding whether a cultural norm is present and adhered to requires correct interpretation of implicit meaning expressed through both nonverbal cues (e.g., gaze, gestures) \citep{pang2024nonverbalculture, hessels2025gaze}, and verbal features (e.g., sarcasm) \citep{rakov13_interspeech, tepperman06b_interspeech}. Additional challenge stems from the cross-cultural aspect, due to the documented differences in norms and values across  cultures \citep{Inglehart_2018, hofstede2009geert, gelfand2011differences}. 

To systematically evaluate cultural norm understanding in videos, we propose a human-AI collaboration framework to collect cultural norm annotations from video clips from comparable US and Chinese TV shows. We use this high-quality dataset to benchmark smaller VideoLLMs cultural competence. We focus on TV shows since sociological research argues %establish 
that scripted television strongly reflects real-world cultural expectations \citep{hawkins1982television, abu1997interpretation, fiske1992audiencing, hawkins1981using}. We select eight popular television shows, four from the US and four from China, representing both formal (workplace) and informal social settings across drama and comedy genres. 

%in videos. 
%First, given a video clip, a powerful VideoLLM generates candidate annotations, extracting an applicable cultural norm and whether norm adherence or violation took place. Second, three human annotators grounded in the relevant culture edit or agree with the candidate annotations.

%We focus on two countries with differing cultural values: US and China, recruiting three monocultural annotators with significant lived experience in each country. As a result, we construct \textsc{VideoNorms}, a dataset of over 3,000 annotations of video clips from $8$ comparable US and Chinese shows. An analysis of annotator disagreements revealed a significant performance degradation of VideoLLM candidate annotations in Chinese cultural context. 

%To benchmark the performance of open-weight models, we introduce \textsc{VideoNorms-Benchmark}, a consensus-filtered high-reliability subset of over $500$ instances where at least $3$ annotators agree. We propose $2$ tasks: (1) binary classification, where given a video clip the model has to predict whether a particular cultural norm was adhered to or violated; (2) an explanation task where the model also has to provide verbal and non-verbal evidence to support its adherence or violation label. In addition, we conduct ablation studies on model size and the need for video modality.

Our contributions are the following:

\begin{itemize}
    \item \textbf{Human-AI collaboration framework for cultural norm annotation in videos.} To construct a dataset of videos annotated for socio-cultural norms,  
    we propose and evaluate a two-stage human-AI collaboration framework based on prior work \cite{chakrabarty-etal-2022-flute, saakyan-etal-2025-understanding} and applied to the video modality: (1) given a 15 second video clip, a large VideoLLM (\texttt{Gemini-2.0-Pro} model \citep{Google2024Gemini}) extracts candidate annotations of socio-cultural norms,  norm categories, adherence/violation labels together with the corresponding %supporting them with relevant 
    verbal and non-verbal evidence (left panel on Figure \ref{fig:pipeline});  (2) 3 trained monocultural annotators with lived experience in the culture review and edit errors in these candidate annotations (right panel on Figure \ref{fig:pipeline}). 
    % \item  \textbf{Analysis of disagreements} between VideoLLM and human annotations, across countries and TV show types, revealing disparate quality of VideoLLM annotations in US and Chinese cultural contexts.
    \item \textbf{\textsc{VideoNorms}}, a dataset of over 3,000 annotations of socio-cultural norms in four pairs of comparable US-Chinese TV shows. In addition, we release \textsc{VideoNorms-Benchmark}, a consensus-based filtered high-reliability subset of $542$ instances. It consists of video clips paired with human-edited/validated annotations of norm category, adherence/violation label, and verbal and non-verbal evidence. 
    \item \textbf{Empirical analysis.} Currently, no studies compare cross-cultural performance of open-weight VideoLLMs w.r.t. understanding of socio-cultural norms. We propose $2$ tasks: (1) binary classification, where given a video clip the model has to predict whether a particular cultural norm was adhered to or violated; (2) an explanation task where the model also has to provide verbal and non-verbal evidence to support its adherence or violation label. Using hierarchical linear modeling, we find that 1) models exhibit performance disparity in US vs. Chinese norm adherence prediction; 2) models have more difficulty in providing non-verbal evidence compared to verbal for norm adherence or violation. Ablations confirm the video modality is necessary for high performance, and scaling model size alone does not yield classification score improvements.
    % Additionally, models struggle with providing evidence for their adherence/violation judgments and identifying the actual norms in videos. These limitations emerge across architectures, suggesting structural gaps in current video-language modeling.
\end{itemize}

\section{Related Work}
\paragraph{Social Norms, Morality, and Cultural Knowledge in NLP}

Textual datasets such as ATOMIC, Social IQa, Social Chemistry 101 modeled everyday commonsense and social interactions \citep{sap2019atomic, sap2019socialiqa, forbes2020socialchem}. Moral Stories added structured narratives to probe norm adherence vs. violation and consequences \citep{emelin2021moralstories}. 
% ETHICS/Delphi framed moral judgments at scale but also prompted critical reflection on dataset design, value pluralism, and the pitfalls of single “gold” labels \citep{hendrycks2021ethics, jiang2025delphi, talat2022delphicritique}. 
NormBank compiled 155k situational norms grounded in roles and settings, moving beyond decontextualized rules \citep{ziems2023normbank}. Complementary cross-cultural resources, GeoMLAMA, Candle, and CulturalBench, probe geographic and cultural knowledge, showing that model performance can vary widely across regions and languages \citep{yin2022geomlama, ponti2020xcopa, nguyen2022ccsk, chiu2024culturalbench}. 
In multimodal contexts, most work has focused on cultural knowledge \citep{winata-etal-2025-worldcuisines, shafique2025culturallydiversemultilingualmultimodalvideo, kadiyala-etal-2025-uncovering} and culturally-aligned generation \citep{nayak2025culturalframesassessingculturalexpectation, havaldar-etal-2025-towards} recently also exploring norm understanding \citep{fung2025normlens, rezaei-etal-2025-egonormia}. Our main contribution is a dataset for cross-cultural social norm understanding requiring fine-grained comprehension of implicit meaning in speech and non-verbal cues across two cultures. 

\paragraph{Video Understanding and Video-Language Models}

Modern VideoLLMs connect powerful vision encoders with LLMs with instruction tuning \citep{alayrac2022flamingo, liu2023llava, lin2024videollava}. However, benchmarks predominantly target temporal or narrative understanding: MVBench, Video-MME, NExT-QA, TGIF-QA, MSRVTT-QA, MSVD-QA, and MovieQA \citep{li2024mvbench, fu2025videomme, xiao2021nextqa, jang2017tgifqa, xu2016msrvtt, tapaswi2016movieqa}. Closer to social reasoning, MovieGraphs annotates human-centric situations (relations, emotions, motivations) in movies \citep{vicol2018moviegraphs}, and \citet{mathur-etal-2025-social} explores social reasoning in videos. To our knowledge, none of these explicitly evaluate cultural norm adherence/violation.

\section{\textsc{VideoNorms} %Dataset and
Dataset} \label{sec:data_construction}

To build \textsc{VideoNorms}, we designed a multi-stage human-AI collaboration framework
that integrates video sampling, AI-assisted norm extraction, and human refinement. Figure~\ref{fig:pipeline} illustrates the overall process. The collaboration framework has 3 steps: %pipeline proceeds in five steps:
(1) select clips from eight US and Chinese television shows spanning both formal and informal contexts (Section \ref{sec:clips}); (2) use a large VideoLLM (Gemini 2.0) to extract candidate norm category, adherence/violation, and verbal/non-verbal evidence (Section \ref{sec:teacherLM}); (3) refine the model-generated data by obtaining edits from $3$ trained annotators with a relevant cultural background on each instance (Section \ref{sec:annot}).

\subsection{Video Selection} \label{sec:clips}

To construct a comparable cross-cultural benchmark for cultural norm recognition, we follow prior work on video QA based on scripted media \citep{tapaswi2016movieqa}. We selected eight popular television shows, four from the US and four from China, representing both formal (workplace) and informal social settings across drama and comedy genres. We turn to TV shows since sociological research establishes that scripted television strongly reflects real-world cultural expectations \citep{hawkins1982television, abu1997interpretation, fiske1992audiencing, hawkins1981using}. While interactions are scripted, they still show behavior that can be judged as adhered to or violating social norms. For the US dataset, workplace shows include \textit{Suits} (drama), depicting professional interactions in a corporate law firm, and \textit{The Office} (comedy), a workplace mockumentary. For informal settings, we chose \textit{Friends} and \textit{The Big Bang Theory}, both portraying casual interactions among close friends. The humorous context allows us to mine more norm violations, since those can occur for comedic effect (e.g., sarcasm). The Chinese dataset mirrors this structure with \textit{The Best Partner} (workplace drama) and \textit{Amazing Night} (workplace comedy) for formal contexts, and \textit{iPartment} and \textit{Home With Kids}, both sitcoms depicting casual interactions among young adults and family members respectively, for informal scenarios. For each show, we sample 15-second clips as input for the video language model (see Appendix \ref{app:clipping}).

\subsection{Candidate annotation} \label{sec:teacherLM}
%\smara{I'll rewrite a bit here to say that socio-cultural norm annotation is a complex task. LDC had guidelines and human only task with very low IAA. Thus studies in this space in txt used human-AI collaboration}
During a pilot annotation experiment, we found that asking the annotators to come up with norm annotations of videos from scratch was highly ineffective: first, there can be many candidate norms, allowing for a lot of disagreement; second, the task would take too much time, becoming too expensive and causing annotator fatigue. Following prior work, we leverage a human-AI collaboration framework that has been shown effective in tasks from figurative meaning interpretation \cite{chakrabarty-etal-2022-flute, saakyan-etal-2025-understanding} to cultural norm discovery \citep{fung-etal-2023-normsage, li-etal-2023-normdial, ch-wang-etal-2023-sociocultural}. Specifically, norms are first generated by a strong multimodal model, and then annotated by human raters for correctness. In this way, our dataset focuses on correctness rather than coverage of all possible norms.
%During a pilot annotation experiment, we found that asking the annotators to come up with norm annotations of videos from scratch was highly ineffective: first, there can be many candidate norms, allowing for a lot of disagreement; second, the task would take too much time, hence becoming too expensive and causing annotator fatigue. Instead, we turn to a human-AI collaboration framework utilized in the past for cultural norm discovery \citep{fung-etal-2023-normsage, li-etal-2023-normdial, ch-wang-etal-2023-sociocultural}.

%\paragraph{Prompting.} 
%\smara{I can re-write this to ground it in LDC}
The Linguistic Data Consortium (LDC) \citep{LDC2022E18} proposes a taxonomy of ten categorizations of social norms of conversations that have been used in prior work on norm discovery and annotations in textual dialogues \citep{li-etal-2023-normdial}. %in its guidelinesOur prompt for cultural norm annotation was inspired by speech act theory \citep{searle1969, austin1962} and based on prior work on norm understanding \citep{li-etal-2023-normdial} and the Linguistic Data Consortium taxonomy \citep{LDC2022E18}. 
% Classic speech act theory: distinguishes between different types of illocutionary acts, identifying five broad classes: \textit{assertives}, \textit{directives}, \textit{commissives}, \textit{expressives}, and \textit{declarations}. 
 Starting from this taxonomy %and prior work 
 and an initial annotation of 50 video clips in a pilot study, we select the following subset of norm categories: %were selected:
 \textit{Thanks}, \textit{Apology}, \textit{Admiration}, \textit{Greeting}, \textit{Farewell}, \textit{Requesting Information}, \textit{Rejecting a Request}, \textit{Granting a Request}, \textit{Expressing criticism}, \textit{Agreement}, and \textit{Disagreement}. 
 % These categories also align with prior work on norm understanding \cite{li-etal-2023-normdial} based on the Linguistic Data Consortium taxonomy \cite{LDC2022E18}. 
 We also include a \textit{Custom Category} which provides flexibility to capture any remaining %multimodal 
 norms not included in the above set. 

To generate candidate annotations of video clips we use \texttt{Gemini-2.0-Pro} model \citep{Google2024Gemini}, as it was the only large video model offering integrated video and audio long-context understanding at the time of the study. The model is prompted to select the most relevant category and generate an applicable cultural norm within that category. An applicable cultural norm is both \textit{relevant to the scene} (i.e., exhibited by the characters in the video clip) and \textit{typical for the culture} (i.e., consistent with expected behavior in that cultural context). For example, in a US business setting, shaking hands during introductions is both relevant to greeting scenes and typical of US professional culture. Our prompt provides examples of norms in both formal and casual contexts for each category (for example, one can apologize with ``Ooops!'' in an informal context, and ``I would like to express my apology'' in a formal context). Besides identifying the norm and the norm category, the model is asked to generate the context in which the situation occurs, the subjects of the norm, whether the norm is adhered to or violated, and provide an explanation with verbal and non-verbal evidence from the clip for the adherence/violation label (see Table \ref{tab:field-mapping}). The prompt was also adapted for Chinese culture, including the translation into Mandarin. See full prompts in Appendix, Table \ref{tab:prompt-comparison-full}.
 \begin{table}[ht]
\small
\centering
\renewcommand{\arraystretch}{0.9} % consistent row height
\begin{tabularx}{\linewidth}{>{\centering\arraybackslash}m{0.18\linewidth} 
                               >{\centering\arraybackslash}m{0.18\linewidth} 
                               >{\raggedright\arraybackslash}X}
\hline
\textbf{English Field} & \textbf{Mandarin Field} & \textbf{Explanation} \\
\hline
Timestamp & 时间戳 & Start and end time of the event. \\
\hline
Context & 情境描述 & Brief description of the setting and the social hierarchy between participants (e.g., colleagues, friends, siblings). \\
\hline
Norm Category & 规范类别 & Selected from the predefined list or dynamically generated under the Custom Category. \\
\hline
Norm Subject & 行为主体 & The individual to whom the norm is applied (without using character names). \\
\hline
Specific Norm & 具体规范 & A precise articulation of the expected behavior for the given context. \\
\hline
Norm Adherence & 规范遵循情况 & Indicates whether the behavior represents adherence or violation. \\
\hline
Explanation & 解释说明 & A breakdown of verbal and nonverbal cues (e.g., dialogue, tone, facial expression, and body language) that justify the assessment. Includes subfields: 语言证据 (verbal evidence) and 非语言证据 (nonverbal evidence). \\
\hline
\end{tabularx}
\caption{English and Mandarin output fields, with explanations.}
\label{tab:field-mapping}
\end{table}

VideoLLM was prompted to produce 511 unique (video clip, norm) candidate annotations across the US dataset and 500 across the CN dataset. 

\subsection{Human annotation} \label{sec:annot}
To collect human annotations of the norms, each candidate item generated above was further inspected by $3$ human annotators. We ensured each annotator's relevant monocultural identity by confirming their country of residence, primary language, monocultural self-identification, earliest language in life, and other factors (see Appendix \ref{sec:screening}). Detailed instructions were provided (see Appendix \ref{app:ann_instrs_ui}), and each annotator was first trained on $20$ instances, where their answers were reviewed to ensure high-quality completion and instruction following. All annotators were fairly compensated above the local minimum wage guidelines.

Annotators validated or modified each field in Table \ref{tab:field-mapping} to verify the selected timestamps, norm category, adherence or violation label, and ensure the verbal and nonverbal evidence correctness. The annotation interface includes an Additional Explanation field where they briefly explain any modifications or confirm agreement with the candidate output. Importantly, \emph{it was necessary to confirm agreement with candidate annotation in case of no edits in order to move on to the next instance}. See a details of the interface in Appendix \ref{app:ann_instrs_ui}. Figure~\ref{fig:annotation-example} compares the large VideoLLM's initial outputs with annotator refinements for both US and Chinese shows, showing extensive involvement of human raters. The annotated data statistics are displayed in Table \ref{tab:dataset-stats}. Note there is a class imbalance between adherence and violation cases, as scripted television predominantly depicts normative behavior, particularly in Chinese shows \citep{zhu2013television}.

\begin{figure*}[htbp]
    \centering
    \includegraphics[width=0.9\textwidth]{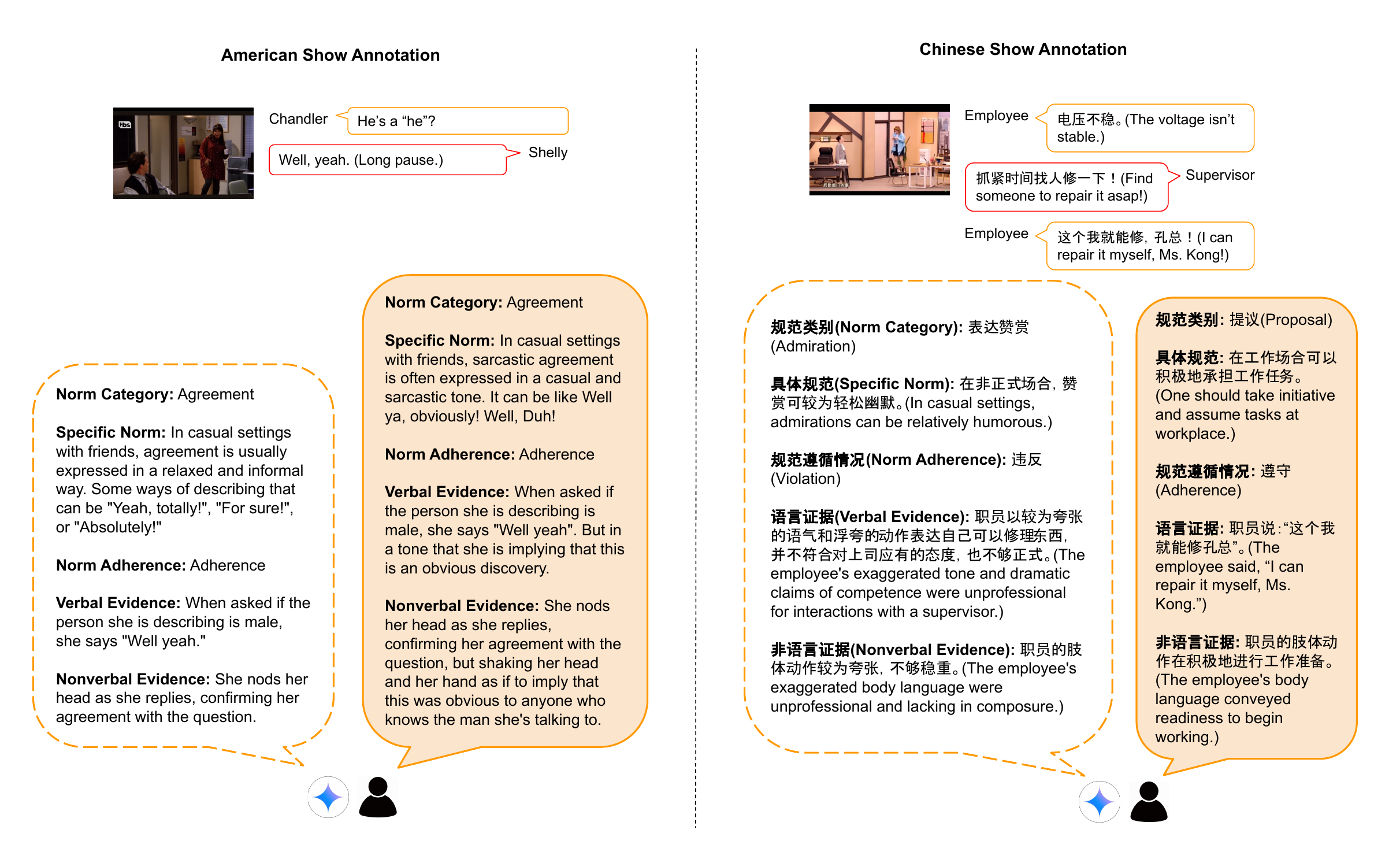}
    \caption{Examples of Gemini-generated normative behavior annotations and corresponding human refinements for US (left) and Chinese (right) shows, as recorded through the annotation interface.}
    \label{fig:annotation-example}
\end{figure*}

\subsection{Agreement and revision rate} \label{sec:disagrement_analysis}
Due to a skewed distribution of norm adherence and violations, traditional agreement metrics would suffer from prevalence bias and would not be suitable \citep{ 10.1162/089120104773633402, brennan1992statistical, FEINSTEIN1990543}.
Hence, we report the free-marginal multirater Kappa $\kappa_{free}$ \citep{randolph2005free} to measure inter-annotator agreement. We report agreement for adherence/violation when the raters agree on the norm category and the social norm. As shown in Table \ref{tab:fleiss-kappa}, agreement for adherence/violation remained relatively high, with both US and Chinese contexts achieving highest agreement for drama shows set in workplace setting (\emph{Suits} $\kappa_{free} = 0.89$, Best Partner $\kappa_{free} = 0.95$). 

For both cultural contexts, substantial amount of candidate annotations was deemed in need of revision. The combined annotator change percentages (total number of fields across all norms and all annotators that have been edited divided by the total number of fields)\footnote{See Appendix \ref{app:changes} for breakdown by field.} for US shows ranged from 18.28\% to 26.04\%. 
In contrast, combined annotator change percentages for Chinese shows were notably higher, ranging from 42.35\% to 53.39\%, \emph{cautioning against fully automated cultural norm extraction}, especially in in under-represented cultures. This aligns with prior findings that LLMs perform better in WEIRD (Western, Educated, Industrialized, Rich, and Democratic) cultural contexts \citep{Mihalcea_Ignat_Bai_Borah_Chiruzzo_Jin_Kwizera_Nwatu_Poria_Solorio_2025, liu-etal-2025-culturally}. 

% Consequently, annotators in the Chinese dataset needed to modify a larger fraction of candidate annotations, which naturally contributes to lower inter-annotator agreement metrics ($0.27-0.45$).

\begin{table}[h]
\centering
\begin{adjustbox}{width=\columnwidth}
\begin{tabular}{l l l l c c}
\hline
\textbf{Dataset} & \textbf{Genre} & \textbf{Setting} & \textbf{Show} & \textbf{Fleiss's $\kappa$} & \textbf{Change \%} \\
\hline
\multirow{4}{*}{US} 
& Comedy & Informal   & Friends & 0.61 & 23.36\% \\
& Comedy & Informal   & Big Bang Theory & 0.71 & 18.28\% \\
& Comedy & Workplace  & The Office & 0.59 & 26.04\% \\
& Drama  & Workplace  & Suits & 0.76 & 20.62\% \\
\hline
\multirow{4}{*}{Chinese} 
& Comedy & Informal   & iPartment &  0.70 & 50.00\% \\
& Comedy & Informal   & Home with Kids & 0.59
 & 46.89\% \\
& Comedy  & Workplace  & Amazing Night & 0.73 & 53.39\% \\
& Drama  & Workplace  & Best Partner & 0.66 & 42.35\% \\
\hline
\end{tabular}
\end{adjustbox}
\caption{Fleiss’s $\kappa$ Scores and Annotator Change Percentages by Genre and Setting.}
\label{tab:fleiss-kappa}
\end{table}

% When comparing across genre (comedy/drama) and setting (workplace/informal), both US and Chinese subsets display higher agreement for workplace drama shows (Suits $\kappa=0.76$ and Best Partner $\kappa = 0.45$), while showing lower agreement on comedy shows. Unlike for US shows, the agreement for Chinese shows in informal settings yielded lower agreement than workplace settings ($\kappa = 0.27-0.33$ vs. $\kappa = 0.42-0.45$), aligning with findings conform with psychology studies Sino-US workplace flexibility \cite{Lai04022022, confvalues}.

\subsection{Consensus-based filtering}
\label{sec:final_stats}
%\smara{I can try to sharpen this, it is too windy. We need to say we take majority vote, and in case of 2-1 we also do another round with 2 meta-annotators.
We will release the full dataset of human annotations for the study of disagreement and pluralistic approaches \citep{Aroyo_Welty_2015, davani-etal-2022-dealing, prabhakaran-etal-2021-releasing}. However, in order to investigate model performance, we opt to select a high-agreement subset of norm annotations. To do this, we select all items in which all three annotators agree on context, category, subject, specific norm, and adherence/violation (allowing multiple (non-)verbal explanations). For items with a $2$ to $1$ split, we additionally recruit $2$ meta-annotators and only then take the $4/5$ majority vote.
This subset, which we refer to as \textsc{VideoNorms-Benchmark}, contains $542$ instances, $273$ in US and $269$ in Chinese cultural context (statistics in Table \ref{tab:dataset-stats}).

\begin{table}[h]
\centering
\small
\begin{adjustbox}{width=0.8\columnwidth}
\setlength{\tabcolsep}{2pt}
\begin{tabular}{lcc}
\toprule
\textbf{Statistic} & \textbf{US} & \textbf{CN} \\
\midrule
\quad Number of Video Clips & 266 & 249 \\
\quad Total Annotated Items & 511 & 500 \\
\quad Total Annotations & 1,533 & 1,500 \\
\midrule
\multicolumn{3}{l}{\textbf{Annotation Agreement}} \\
\quad All 3 Annotators Agree & 54.0\% & 14.2\% \\
\quad 2 Agree, 1 Differs & 37.8\% & 45.0\% \\
\quad All 3 Disagree & 8.2\% & 40.8\% \\
\midrule
\multicolumn{3}{l}{\textbf{Benchmark Dataset}} \\
\quad Total Samples & 273 & 269 \\
\quad Adherence & 63.7\% & 82.9\% \\
\quad Violation & 36.3\% & 17.1\% \\
\midrule
\multicolumn{3}{l}{\textbf{Top Norm Categories}} \\
\quad Expressing criticism / 提出批评 & 33.7\% & 7.1\% \\
\quad Requesting information / 请求信息 & 24.2\% & 17.8\% \\
\quad Admiration / 表达赞赏 & 8.8\% & 8.6\% \\
\quad Greeting / 问候 & 8.8\% & 5.6\% \\
\quad Rejecting a request / 拒绝请求 & 3.3\% & 3.3\% \\
\bottomrule
\end{tabular}
\end{adjustbox}
\caption{Annotation agreement and benchmark dataset statistics for English (US) and Chinese (CN) norms. Agreement is computed over all annotated items prior to filtering; final dataset reflects majority-vote filtered samples. Category percentages show share of total samples per language.}
\label{tab:dataset-stats}
\end{table}

\section{Experiments}
\label{ref:experiments}
\paragraph{Tasks.} We evaluate cultural norm understanding with the following two tasks. 

\emph{Task 1: adherence or violation prediction}.  
Given a video segment, transcript (obtained with \texttt{whisper-large-v3} \citep{radford2022whisper}), norm category, and a specific norm, predict whether the behavior observed in the video \emph{adheres to} or \emph{violates} that norm (binary classification). 

\emph{Task 2: explainable adherence or violation prediction}.  
Building on Task~1, the model must (i) predict Adherence/Violation and (ii) generate (a) verbal evidence referencing the spoken content (e.g., quoted phrases), and (b) nonverbal evidence referencing visual social cues (e.g., gaze, gesture, distance, facial expression). 

% To obtain evidence quality scores, we compare generated evidence to the reference evidences in the dataset adopting an LLM-as-judge (LLM-J) protocol. GPT-5 grader compares verbal/nonverbal rationales for correctly predicted labels to the reference evidences in the benchmark using a 5-point rubric. \footnote{When multiple evidences are present due to annotator edits, we take the maximum similarity score among them.}
% Our rubric explicitly distinguishes between content perception (scores 1-2: missing visual/verbal cues; score 3: cues identified without reasoning) and cultural reasoning (scores 4-5: culturally-appropriate interpretation matching human annotations), see examples in Table \ref{tab:rubric_examples}. The LLM-J provides a rationale plus numeric verdict for \emph{Verbal} and \emph{Nonverbal} evidence. See details and validation in Appendix \ref{app:llmJudge}.

\paragraph{VideoLLMs.} We benchmark recent commonly used open-weight VideoLLMs: 
LLaVA-family models (\texttt{LLaVA-Next-Video} \citep{zhang2024llavanextvideo}, \texttt{LLaVA-OneVision} \citep{li2024llavaonevisioneasyvisualtask}) extend image-text pretraining to video by sampling frames and using linear scaling with Rotary Position Embeddings \citep{su2023roformerenhancedtransformerrotary} to achieve long-context understanding. \texttt{InternVL-3} \citep{zhu2025internvl3exploringadvancedtraining} and \texttt{InternVL-3.5} \citep{wang2025internvl35advancingopensourcemultimodal} use variable visual position encoding \citep{ge2024v2peimprovingmultimodallongcontext} for longer multimodal contexts as well as a native multimodal pre-training approach. \texttt{Qwen2-VL }\citep{wang2024qwen2vl} and \texttt{Qwen2.5-VL }\citep{qwen2025qwen25technicalreport} employ dedicated  Multimodal Rotary Position Embeddings (M-RoPE)  to represent temporal and spatial information, enabling the model to comprehend dynamic video content. \texttt{VideoChatR1 }\citep{li2025videochatr1enhancingspatiotemporalperception} uses Reinforcement Fine-Tuning (RFT) with GRPO \citep{shao2024deepseekmath} for video MLLMs to achieve state-of-the-art performance on spatio-temporal perception tasks.
% \begin{itemize}
%     \item LLaVA-family models (\texttt{LLaVA-Next-Video} \citep{zhang2024llavanextvideo}, \texttt{LLaVA-OneVision} \citep{li2024llavaonevisioneasyvisualtask}) extend image-text pretraining to video by sampling frames and using linear scaling with Rotary Position Embeddings \cite{su2023roformerenhancedtransformerrotary} to achieve long-context understanding.
%     \item \texttt{InternVL-3} \citep{zhu2025internvl3exploringadvancedtraining} and \texttt{InternVL-3.5} \citep{wang2025internvl35advancingopensourcemultimodal} use variable visual position encoding \cite{ge2024v2peimprovingmultimodallongcontext} for longer multimodal contexts as well as a native multimodal pre-training approach.
%     \item \texttt{Qwen2-VL }\citep{wang2024qwen2vl} and \texttt{Qwen2.5-VL }\citep{qwen2025qwen25technicalreport} employ dedicated  Multimodal Rotary Position Embeddings (M-RoPE)  to represent temporal and spatial information, enabling the model to comprehend dynamic video content.
%     \item \texttt{VideoChatR1 }\citep{li2025videochatr1enhancingspatiotemporalperception} uses Reinforcement Fine-Tuning (RFT) with GRPO \cite{shao2024deepseekmath} for video MLLMs to achieve state-of-the-art performance on spatio-temporal perception tasks.
% \end{itemize}

\paragraph{Metrics.} 
We report traditional F1 scores for both Task 1 and 2 in Table \ref{tab:mainresults}. All scores are reported excluding items for which models did not follow formatting instructions.

For Task 2, we report evidence quality for correct instances, to disentangle reasoning quality from prediction accuracy, which was evaluated in Task 1. To obtain evidence quality scores, we compare generated evidence to the reference evidences in the dataset adopting an LLM-as-judge (LLM-J) protocol. GPT-5 grader compares verbal/nonverbal rationales for correctly predicted labels to the reference evidences in the benchmark using a 5-point rubric.\footnote{In case of multiple reference (non-)verbal explanations, we take the maximum LLM-J quality score.}
Our rubric explicitly distinguishes between content perception (scores 1-2: missing visual/verbal cues; score 3: cues identified without reasoning) and cultural reasoning (scores 4-5: culturally-appropriate interpretation matching human annotations), see examples in Table \ref{tab:rubric_examples}. The LLM-J provides a rationale plus numeric verdict for \emph{Verbal} and \emph{Nonverbal} evidence. See details and validation in Appendix \ref{app:llmJudge}.

In addition, given a large number of confounding factors (different models, cultural contexts, shows, each item's difficulty and difficulty of adherence or violation label), we turn to hierarchical linear modeling \citep{gelman2007data} for a more robust analysis of model performance \citep{lampinen-etal-2022-language}. We fit a hierarchical logistic regression predicting whether the VideoLLM $j$ was correct on item $i$. The fixed effect input variables were VideoLLM $j$ and control variables for item's cultural context (US vs. China), adherence or violation, and the show it came from. In addition, we control for item and general norm category variation by including random intercepts (see details in Appendix \ref{app:linmodel}). Due to a large number of custom norm categories, to facilitate the linear modeling analysis we group all categories into $4$ general clusters according to \citet{allan1994indirect, allan1998meaning}: Statements (e.g., Agreement), Invitationals (e.g., Requesting Information), Authoritatives (e.g., Granting a Request), Expressives (e.g., Greetings); see Appendix \ref{app:norm_category_clustering})..

\subsection{Results and findings}

\paragraph{Models have different strengths.} As can be seen in Table \ref{tab:mainresults}, Intern3.5-VL slightly outperforms on US classification tasks. Llava-OneVision excels on CN classification, but lags behind VideoChatR1 on Task 2, as the reasoning model preserves classification performance while generating explanations. For generation tasks, Intern3-VL achieves the highest verbal evidence scores (on correct predictions) benefiting from its fine-grained visual grounding. Some models also exhibit performance degradation when the evidence generation requirement is introduced (e.g., Macro F1 of 67.8 in Task 2 vs. 71.5 in Task 1 for Intern3.5-VL), consistent with prior studies on textual explanations \citep{camburu2018snli}. Overall, no single model is a clear leader in all tasks.

\begin{table}[htbp]
\centering
\begin{adjustbox}{width=\columnwidth}
\begin{tabular}{l c c c c}
\toprule
 & \textbf{Task 1} & \multicolumn{3}{c}{\textbf{Task 2}} \\
\cmidrule(lr){2-2}\cmidrule(lr){3-5}
\textbf{Models} & \textbf{Macro F1} & \textbf{Macro F1} & \textbf{Verbal} & \textbf{Nonverbal} \\
\midrule
\multicolumn{5}{c}{\textbf{US Norms}}\\
\midrule
Llava-Next-Video & 39.0$_{\pm2.3}$          & 60.8$_{\pm6.1}$          & 2.37$_{\pm0.17}$          & 2.14$_{\pm0.18}$ \\
Llava-OneVision  & 63.8$_{\pm6.4}$          & 67.4$_{\pm6.1}$          & 2.82$_{\pm0.17}$          & 2.18$_{\pm0.15}$ \\
Intern3-VL       & 57.7$_{\pm5.9}$          & 56.6$_{\pm6.0}$          & \textbf{3.08}$_{\pm0.16}$ & \textbf{2.41}$_{\pm0.19}$ \\
Intern3.5-VL     & \textbf{71.5}$_{\pm5.4}$ & \textbf{67.8}$_{\pm5.6}$ & 2.93$_{\pm0.18}$          & 2.20$_{\pm0.17}$ \\
Qwen2-VL         & 68.5$_{\pm5.7}$          & 62.6$_{\pm6.2}$          & 2.99$_{\pm0.16}$          & 2.15$_{\pm0.17}$ \\
Qwen2.5-VL       & 57.4$_{\pm5.8}$          & 49.4$_{\pm5.9}$          & 2.83$_{\pm0.19}$          & 2.10$_{\pm0.20}$ \\
VideoChatR1      & 69.4$_{\pm5.3}$          & 67.3$_{\pm5.8}$          & 2.86$_{\pm0.17}$          & 2.18$_{\pm0.18}$ \\
\midrule
\multicolumn{5}{c}{\textbf{Chinese (CN) Norms}}\\
\midrule
Llava-Next-Video & 45.3$_{\pm1.5}$          & 50.0$_{\pm6.1}$          & 1.96$_{\pm0.16}$          & 1.64$_{\pm0.14}$ \\
Llava-OneVision  & \textbf{64.9}$_{\pm6.1}$ & 58.2$_{\pm6.0}$          & 2.78$_{\pm0.19}$          & 2.23$_{\pm0.18}$ \\
Intern3-VL       & 37.5$_{\pm5.9}$          & 33.9$_{\pm5.9}$          & \textbf{3.26}$_{\pm0.23}$ & \textbf{2.63}$_{\pm0.23}$ \\
Intern3.5-VL     & 56.4$_{\pm6.0}$          & 58.2$_{\pm6.4}$          & 3.06$_{\pm0.19}$          & 2.61$_{\pm0.18}$ \\
Qwen2-VL         & 61.4$_{\pm6.3}$          & 56.0$_{\pm6.1}$          & 2.80$_{\pm0.20}$          & 2.31$_{\pm0.18}$ \\
Qwen2.5-VL       & 43.3$_{\pm6.0}$          & 42.3$_{\pm6.2}$          & 3.09$_{\pm0.23}$          & 2.29$_{\pm0.22}$ \\
VideoChatR1      & 58.8$_{\pm6.3}$          & \textbf{59.2}$_{\pm6.3}$ & 2.96$_{\pm0.18}$          & 2.29$_{\pm0.18}$ \\
\bottomrule
\end{tabular}
\end{adjustbox}
\caption{Task 1 and Task 2 results for US and Chinese cultural norms. Macro F1 scores shown with $\pm$ variance; highest value per column per culture bolded.}
\label{tab:mainresults}
\end{table}

% \begin{figure}[htbp]
%     \centering
%     \includegraphics[width=\columnwidth]{ figs/a_vs_v.png}
%     \caption{\emph{Adherence vs. Violation Performance.} Comparison of estimated marginal mean probability (averaged over shows and models) of correct prediction for adherence and violation class, within each cultural context. OR = odds ratio.}
%     \label{fig:adherence_vs_violation}
% \end{figure}

% \paragraph{Norm adherence identification is harder than violation in Chinese cultural context.} Figure \ref{fig:adherence_vs_violation} shows how models perform on predicting adherence vs. violation of norms. Hierarchical linear model estimates that across VideoLLMs and shows, the odds of predicting a norm violation correctly are $77\%$ higher than adherence within the Chinese context (OR$=1.77$, $p=0.037$). Within the US context, the difference was not statistically significant $p = 0.119$.

\begin{figure}[h]
    \centering
    \includegraphics[width=\columnwidth]{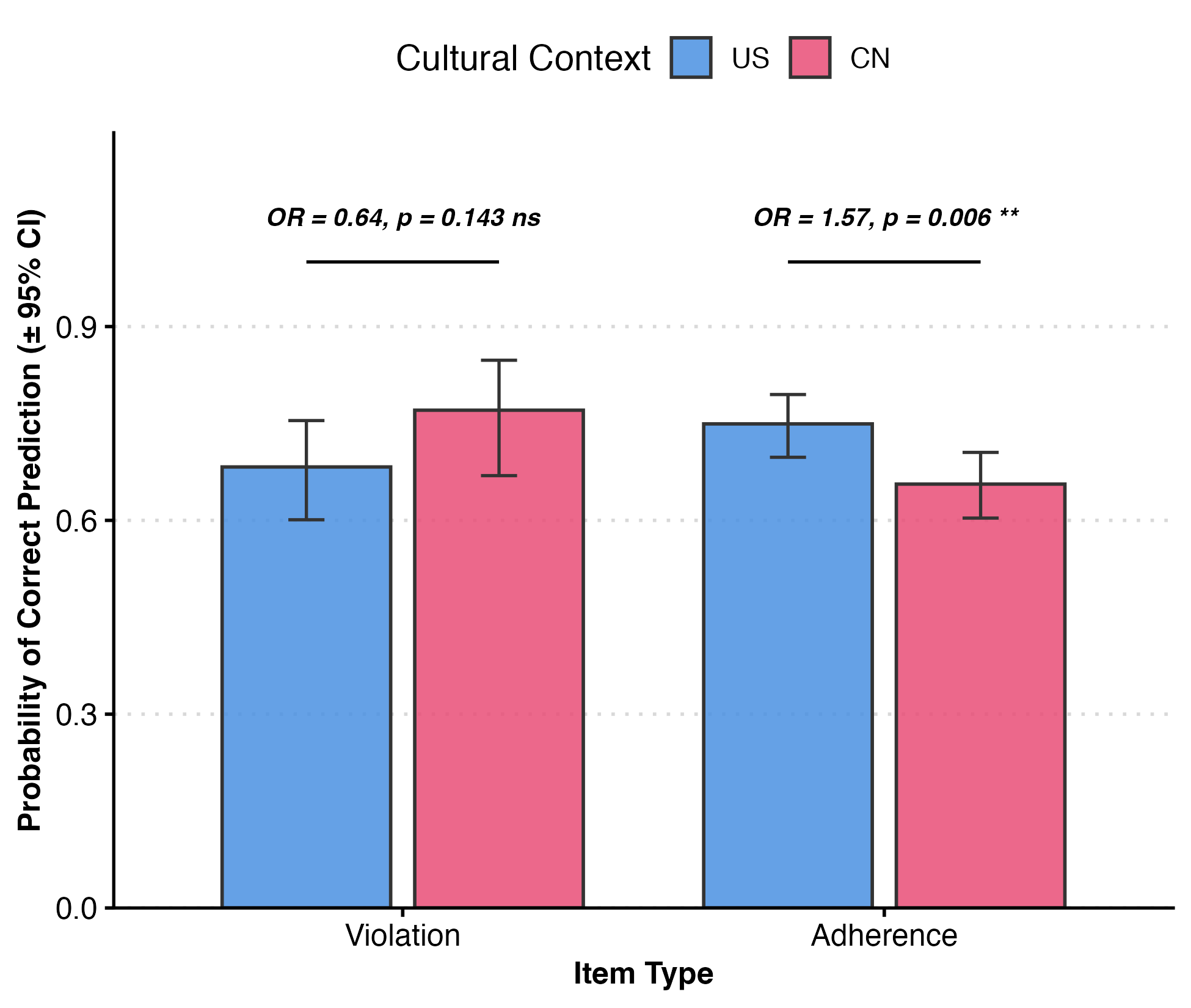}
    \caption{\emph{US vs. CN Performance.} Comparison of estimated marginal mean probability (averaged over shows and models) of correct prediction by cultural context, for adherence and violation classes. OR = odds ratio.}
    \label{fig:bias}
\end{figure}

\paragraph{Models exhibit cultural performance disparity in adherence prediction.}
Figure \ref{fig:bias} shows how models perform in differing cultural contexts. Hierarchical linear model estimates that across VideoLLMs and shows, for norm adherence identification, the odds of correct prediction are $57\%$ higher for US than China (OR$=1.57$, $p=0.006$), with no evidence identified for this effect in violation prediction (possibly due to lower sample size of violations). This suggests that norm adherence prediction requires culturally-specific reasoning.

\begin{figure}[htbp]
    \centering
    \includegraphics[width=\columnwidth]{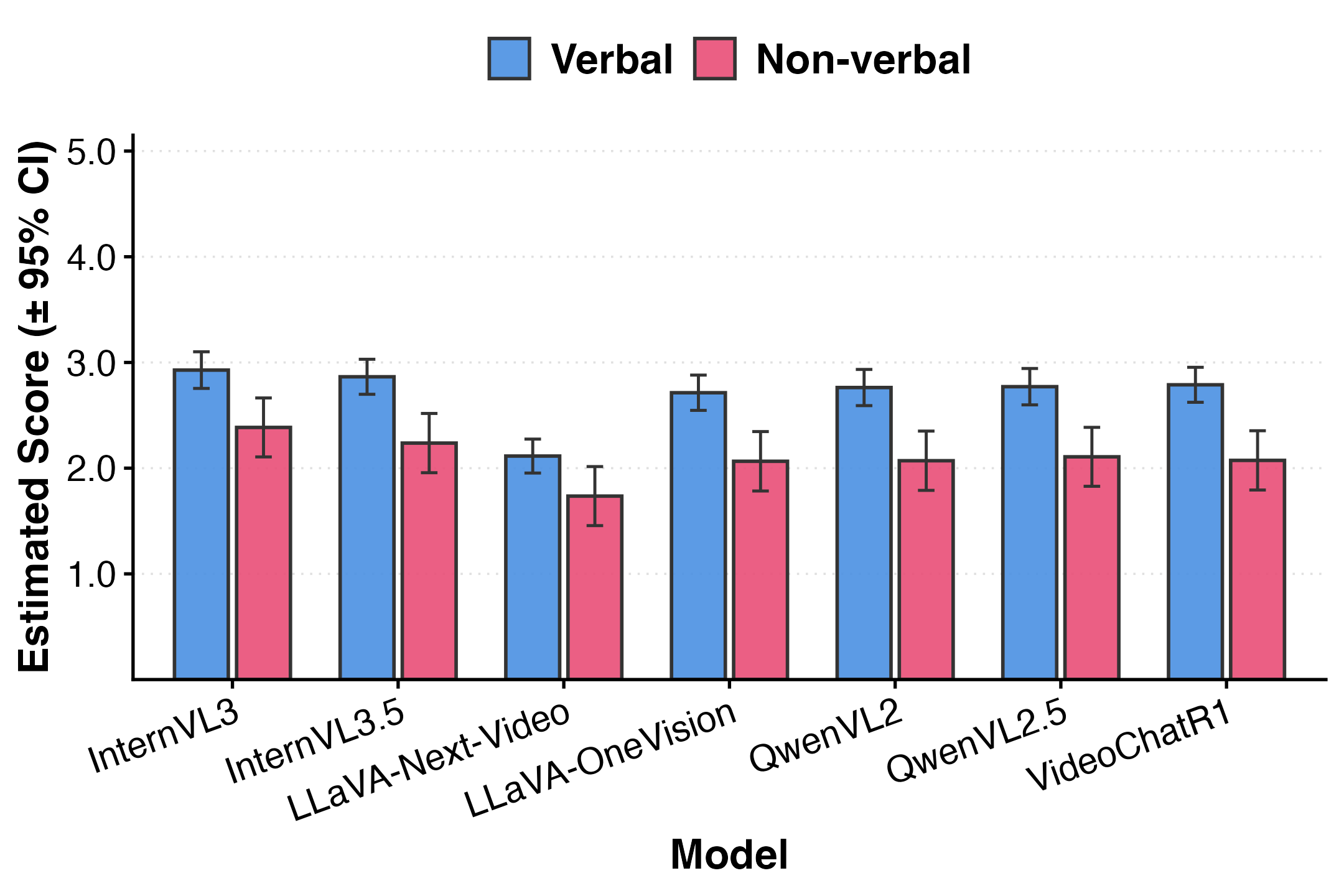}
    \caption{\emph{Verbal and Non-verbal evidence score comparison across models.} Comparison of estimated marginal mean verbal and non-verbal scores (averaged over shows and models).}
    \label{fig:verbal_nonverbval}
\end{figure}

\paragraph{Models struggle with non-verbal evidence more than verbal evidence.} We estimate VideoLLM performance on verbal and non-verbal evidence generation using a hierarchical linear regression model (see Appendix \ref{app:evidence_linmodel}). As shown in Figure \ref{fig:verbal_nonverbval}, models struggle with non-verbal evidence generation, obtaining significantly lower scores. This indicates that while language model backbone is strong enough to analyze the transcript, the multimodal capabilities are not yet on the same level.

\section{Ablation}
We conduct ablation studies using Intern3.5-VL to isolate the contributions of input modality, model size, and number of sampled frames to task performance.

\paragraph{Video modality is necessary for high performance.}
Table~\ref{tab:ablation_modality} compares transcript-only, video-only, and video + transcript performance of Intern3.5-VL 8B. Both cultural contexts' highest classification performance includes the video modality. Video-only achieves the strongest classification performance for Chinese, suggesting a negative effect of transcript addition, while video + transcript performs best for US. The benefit of video over transcript-only is statistically significant for both cultures (p < 0.01 for US, p < 0.001 for CN), and the addition of transcript to video significantly improves US performance (p < 0.001) but not Chinese. This could be (speculatively) explained by two factors: lower quality of transcription models for Chinese and cross-cultural communication differences. For example, according to \citep{hall1976beyond}, US is a low-context culture, where meaning is conveyed primarily through explicit verbal content; by contrast, China is a high-context culture, where meaning is conveyed more through nonverbal cues such as gesture, gaze, and posture. 
% We base our analysis upon Hall's high-context versus low-context cultural communication framework \citep{hall1976beyond}. US is a low-context culture, where meaning is conveyed primarily through explicit verbal content \citep{hall1976beyond}. This may explain why  Chinese, by contrast, is a high-context culture, where meaning is conveyed more through nonverbal cues such as gesture, gaze, and posture \citep{hall1976beyond, hofstede2009geert}. Consistent with this, 
% Video + transcript consistently achieves the highest nonverbal evidence scores across both cultures, confirming the complementary nature of both modalities for evidence generation. 
% Overall, these results demonstrate that video is a valuable input for cultural norm understanding, and that text-based inference alone is not sufficient for high task performance.

\paragraph{Scaling model size does not reliably improve classification performance.}
Table~\ref{tab:ablation_modelsize} compares 4B, 8B, and 14B variants of Intern3.5-VL. For both US and CN contexts, the 8B model achieves the best F1 score and the 14B model leads on evidence quality as expected from a better language model backbone, though confidence intervals overlap. This variable pattern is consistent with prior findings that scaling alone does not resolve culturally grounded reasoning deficits, suggesting the bottleneck lies in culturally informed training data rather than raw model capacity \citep{shi-etal-2024-culturebank, liu-etal-2025-culturally}.

%% ============================================================
%% TABLE A: Input Modality Ablation (EN + ZH)
%% ============================================================
\begin{table}[t]
\centering
\begin{adjustbox}{width=\columnwidth}
\begin{tabular}{l c c c c}
\toprule
 & \textbf{Task 1} & \multicolumn{3}{c}{\textbf{Task 2}} \\
\cmidrule(lr){2-2}\cmidrule(lr){3-5}
\textbf{Input Modality} & \textbf{Macro F1} & \textbf{Macro F1} & \textbf{Verbal} & \textbf{Nonverbal} \\
\midrule
\multicolumn{5}{c}{\textbf{US Norms}} \\
\midrule
Transcript only    & 69.4$_{\pm5.5}$ & 63.6$_{\pm5.7}$ & \textbf{2.99}$_{\pm0.17}$ & 1.79$_{\pm0.16}$ \\
Video only         & 55.6$_{\pm5.9}$ & 47.4$_{\pm5.9}$ & 1.70$_{\pm0.18}$ & \textbf{2.23}$_{\pm0.22}$ \\
Video + Transcript & \textbf{71.5}$_{\pm5.4}$ & \textbf{67.8}$_{\pm5.6}$ & 2.93$_{\pm0.18}$ & 2.20$_{\pm0.17}$ \\
\midrule
\multicolumn{5}{c}{\textbf{Chinese (CN) Norms}} \\
\midrule
Transcript only    & 42.5$_{\pm6.0}$ & 41.5$_{\pm6.0}$ & 3.04$_{\pm0.20}$ & 1.17$_{\pm0.12}$ \\
Video only         & \textbf{61.5}$_{\pm6.0}$ & \textbf{61.0}$_{\pm6.4}$ & 2.41$_{\pm0.17}$ & 2.46$_{\pm0.17}$ \\
Video + Transcript & 56.4$_{\pm6.0}$ & 57.6$_{\pm6.3}$ & \textbf{3.06}$_{\pm0.19}$ & \textbf{2.61}$_{\pm0.18}$ \\
\bottomrule
\end{tabular}
\end{adjustbox}
\caption{Ablation over input modality (Intern3.5-VL 8B)}
\label{tab:ablation_modality}
\end{table}

%% ============================================================
%% TABLE B: Model Size Ablation (EN + ZH)
%% ============================================================
\begin{table}[t]
\centering
\begin{adjustbox}{width=\columnwidth}
\begin{tabular}{l c c c c}
\toprule
 & \textbf{Task 1} & \multicolumn{3}{c}{\textbf{Task 2}} \\
\cmidrule(lr){2-2}\cmidrule(lr){3-5}
\textbf{Model Size} & \textbf{Macro F1} & \textbf{Macro F1} & \textbf{Verbal} & \textbf{Nonverbal} \\
\midrule
\multicolumn{5}{c}{\textbf{US Norms}} \\
\midrule
4B  & 64.1$_{\pm5.5}$ & 60.8$_{\pm5.5}$ & 2.89$_{\pm0.16}$ & 2.24$_{\pm0.19}$ \\
8B  & \textbf{71.5}$_{\pm5.4}$ & \textbf{67.8}$_{\pm5.6}$ & 2.93$_{\pm0.18}$ & 2.20$_{\pm0.17}$ \\
14B & 69.7$_{\pm5.5}$ & 65.4$_{\pm5.6}$ & \textbf{3.02}$_{\pm0.16}$ & \textbf{2.37}$_{\pm0.18}$ \\
\midrule
\multicolumn{5}{c}{\textbf{Chinese (CN) Norms}} \\
\midrule
4B  & 52.1$_{\pm5.9}$ & 50.7$_{\pm5.9}$ & 2.91$_{\pm0.22}$ & 2.51$_{\pm0.17}$ \\
8B  & \textbf{56.4}$_{\pm6.0}$ & \textbf{57.6}$_{\pm6.3}$ & 3.06$_{\pm0.19}$ & 2.61$_{\pm0.18}$ \\
14B & 54.3$_{\pm5.8}$ & 50.8$_{\pm5.8}$ & \textbf{3.08}$_{\pm0.20}$ & \textbf{2.84}$_{\pm0.19}$ \\
\bottomrule
\end{tabular}
\end{adjustbox}
\caption{Ablation over model size (Intern3.5-VL, Video + Transcript).}
\label{tab:ablation_modelsize}
\end{table}
\section{Conclusion}
To evaluate cross-cultural competence in VideoLLMs, we propose a human-AI collaboration framework to collect a dataset of over 3,000 annotations of socio-cultural norms in videos. We focus on two countries with differing cultural values: US and China. Analysis of annotator disagreements revealed disparate performance of candidate VideoLLM annotations between Chinese and US cultural contexts. To evaluate cultural norm understanding of open-weight VideoLLMs, we introduce a benchmark \textsc{VideoNorms-Benchmark}, consisting of a subset where at least $3$ annotators agree. We benchmark a variety of open-source VideoLLMs and find the following trends: 1) models exhibit cultural performance disparity in norm adherence; 2) models have more difficulty in providing non-verbal evidence compared to verbal for norm adherence or violation. Ablation studies confirm video modality is indeed necessary for high task performance, and scaling model size does not yield improvements, indicating the need for inclusion of more diverse cultural data in video model training. Overall, we hope our research serves as an important step towards understanding the cultural competence of video models.

\section{Limitations}
Cultural norm understanding is a complex, nuanced topic requiring many pragmatic simplifications, which can also be observed in prior work \citep{huang-yang-2023-culturally, ch-wang-etal-2023-sociocultural, li-etal-2023-normdial}. First, our work does not claim to measure overall cultural competence, as only two countries were investigated. Second, our work is in no way meant to capture the full cultural diversity the countries represented. Due to lack of data, we had to treat countries like US and China as monolithic despite large intra-cultural variation \citep{plepi2022unifying, wan2023everyonesvoice}, similarly to the ``country as a culture'' fallacy common in our field \citep{alkhamissi-etal-2026-hire}. Overall, our work should be viewed as a practical starting point for cross-cultural norm benchmarking in videos, which has not been done before.

There were also numerous necessary simplifications w.r.t data collection. We had to rely on scripted TV shows as ``cultural mirrors'' \citep{hawkins1981using} instead of naturally occurring interactions between humans to obtain video data. These shows may have occurred in pre-training, though it is unlikely that prior knowledge would affect norm predictions: the task requires fine-grained multimodal reasoning over specific 15-second clips; in addition, prior work \citep{tapaswi2016movieqa} has shown that familiarity does not substitute for task-required reasoning. We could only recruit $3$ annotators per country to annotate a large number of items within the annotation budget; we do not claim that this is representative of such heterogeneous countries as US and China, though we ensured their monocultural identity (see Appendix \ref{sec:screening}). Unavoidably due to the necessity of the human-AI collaboration framework, the benchmark might contain anchor bias due to initial generation by the VideoLLM. To mitigate this, each item was inspected by $3$ annotators independently, resulting in high edit rates.

Finally, we would like to caution against using our benchmark for stereotyping, following the ecological fallacy \citep{brewer2014ecological}: incorrect inference of individual-level traits based on aggregated national-level culture data. In addition, as with any benchmark, there is a potential risk of over-optimization to the metric, where high performance does not necessarily mean that the model is culturally competent. We hope that our dataset paves the way for larger-scale, more representative, longitudinal annotations of socio-cultural norms using our framework.

% Bibliography entries for the entire Anthology, followed by custom entries
%\bibliography{anthology,custom}
% Custom bibliography entries only
\bibliography{custom}
\clearpage
\newpage
\appendix

\section{Candidate Data Details}

\subsection{Video Clipping and Segmentation} \label{app:clipping}

We collected 5-7 clips of 2-3 minutes each from YouTube for every TV show in our dataset. YouTube clips were selected instead of full episodes because they are easier to download and process into smaller segments. After collection, a Python script was used to divide each 2-3 minute clip into 15-second sub-clips. This length was chosen because a 15-second segment typically contains one distinct social norm, allowing the model to focus on a single interaction or event. Shorter segments also help Gemini generate more precise and detailed norm outputs, which are described in the following section.

\subsection{Prompting details}
Table \ref{tab:prompt-comparison-full} shows the full speech act theory-based prompt for candidate cultural norm annotation.

% Reset spacing if needed
\setlength{\LTpre}{\bigskipamount}
% Reset spacing if needed
\setlength{\LTpre}{\bigskipamount}
% Reset spacing if needed
\setlength{\LTpre}{\bigskipamount}

\subsection{Annotator Change Percentages by Field} \label{app:changes}

Table \ref{tab:field-change-percentages} shows percentages of changes per field by U.S. and Chinese annotators.

\begin{table}[h]
\centering
\small
\begin{adjustbox}{width=0.98\columnwidth}
\begin{tabular}{lcc}
\hline
\textbf{Field} & \textbf{US (\%)} & \textbf{China (\%)} \\
\hline
timestampStart & 2.40 & 4.86 \\
timestampEnd & 3.05 & 11.78 \\
context & 4.99 & 56.82 \\
normCategory & 14.59 & 50.83 \\
normActors & 29.44 & 53.83 \\
specificNorm & 16.99 & 57.88 \\
normAdherence & 13.62 & 26.28 \\
verbalEvidence & 23.35 & 64.54 \\
nonverbalEvidence & 19.71 & 53.89 \\
% additionalFeedback & 97.60 & 99.93 \\
\hline
\end{tabular}
\end{adjustbox}
\caption{Overall Percentage of Changes per Field by U.S. and Chinese Annotators}
\label{tab:field-change-percentages}
\end{table}

\section{Annotation Details}

\subsection{Screening Questions and Answers} \label{sec:screening}

Below are the screening questions used for both American and Chinese annotators:

\begin{itemize}
    \item Were you raised monolingual?
    \item What's your earliest language in life, and what's your primary language?
    \item Do you identify yourself as monocultural or multicultural?
    \item What's your highest education level completed?
    \item What is your country of birth? How many years have you lived in your current country of residence?
\end{itemize}

For the U.S. annotators, all three were raised monolingual. English is the first and primary language of all three annotators. All three identify as monocultural and have completed a Bachelor's degree. All were born in the United States and have lived there for their entire lives.

For the Chinese annotators, all three were raised monolingual. Mandarin is the first and primary language of all three annotators. All three identify as monocultural and have completed a Bachelor's degree. All were born in China and have lived there for their entire lives.
% \subsection{Annotator Survey}
% After the annotators received their offers, we asked them to complete a short follow-up survey. This survey included questions about their demographic background and prior exposure to the shows used in the study. Specifically, annotators were asked to provide their age, indicate which of the four shows they had previously watched, and rate their familiarity with each show.
The U.S. annotators were 27, 28, and 29 years old, while the Chinese annotators were between 25, 28, and 29 years old.

% For the U.S. annotators, two had watched \textit{Friends} and \textit{Big Bang Theory}, while all three had watched \textit{The Office} and \textit{Suits}. Of the two annotators who had watched \textit{Friends}, one gave a familiarity rating of $5/5$ and the other gave a $2/5$. For \textit{Big Bang Theory}, both annotators gave a familiarity rating of $5/5$. For \textit{The Office}, two annotators gave a familiarity rating of $5/5$, and the third gave a $4/5$. For \textit{Suits}, two annotators gave a familiarity rating of $5/5$, and the third gave a $3/5$.\\

% For the Chinese annotators, all three had watched \textit{iPartment} and \textit{Home with Kids}, while only one had watched \textit{Amazing Night} and none had watched \textit{Best Partner}. For \textit{iPartment}, two annotators gave a familiarity rating of $5/5$, and the third gave a $4/5$. For \textit{Home with Kids}, the first annotator gave a familiarity rating of $5/5$, the second gave a $4/5$, and the third gave a $3/5$. The annotator who had watched \textit{Amazing Night} gave a familiarity rating of $3/5$. 

\subsection{Annotation Instructions and User Interface} \label{app:ann_instrs_ui}

\begin{figure*}[h]
\centering
\includegraphics[width=0.3\textwidth]{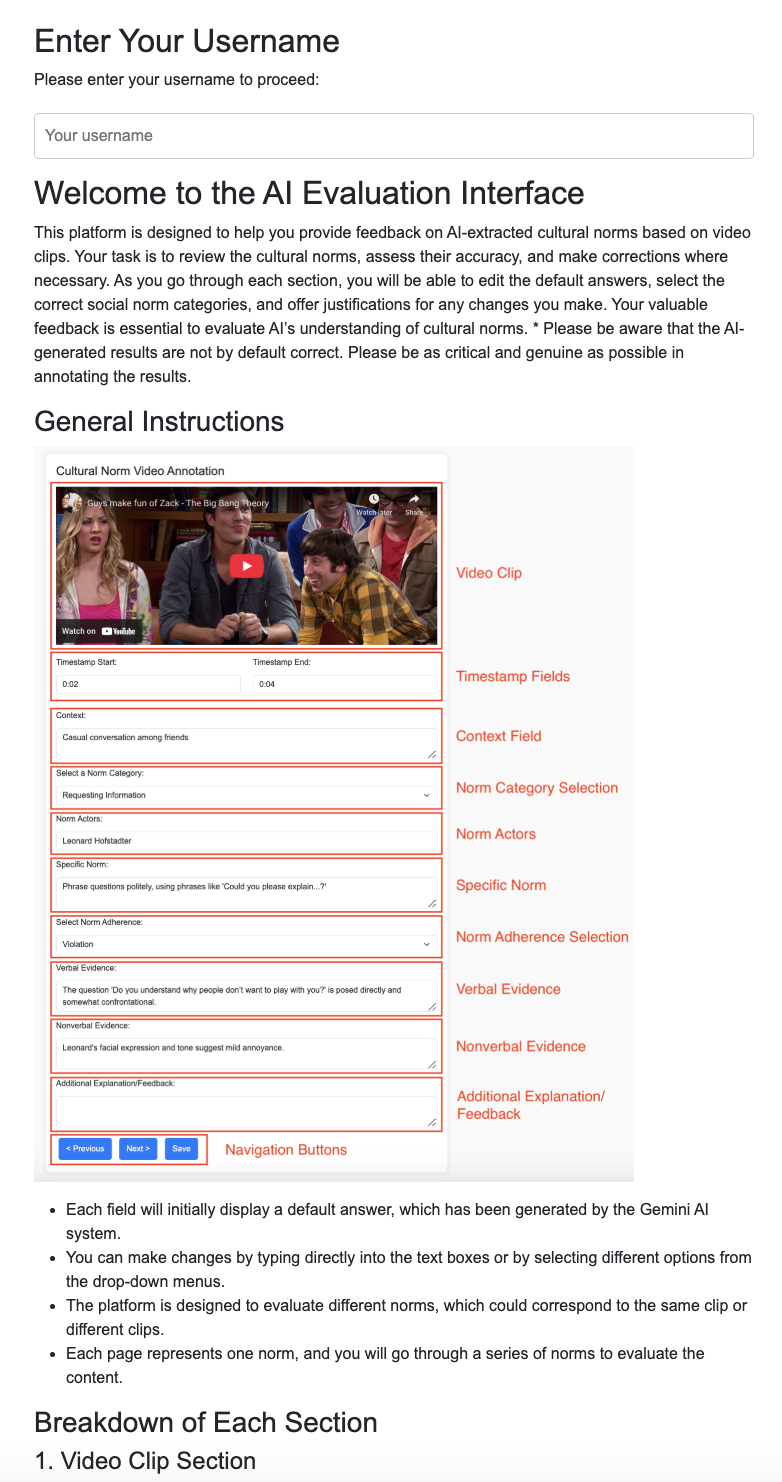}
\includegraphics[width=0.3\textwidth]{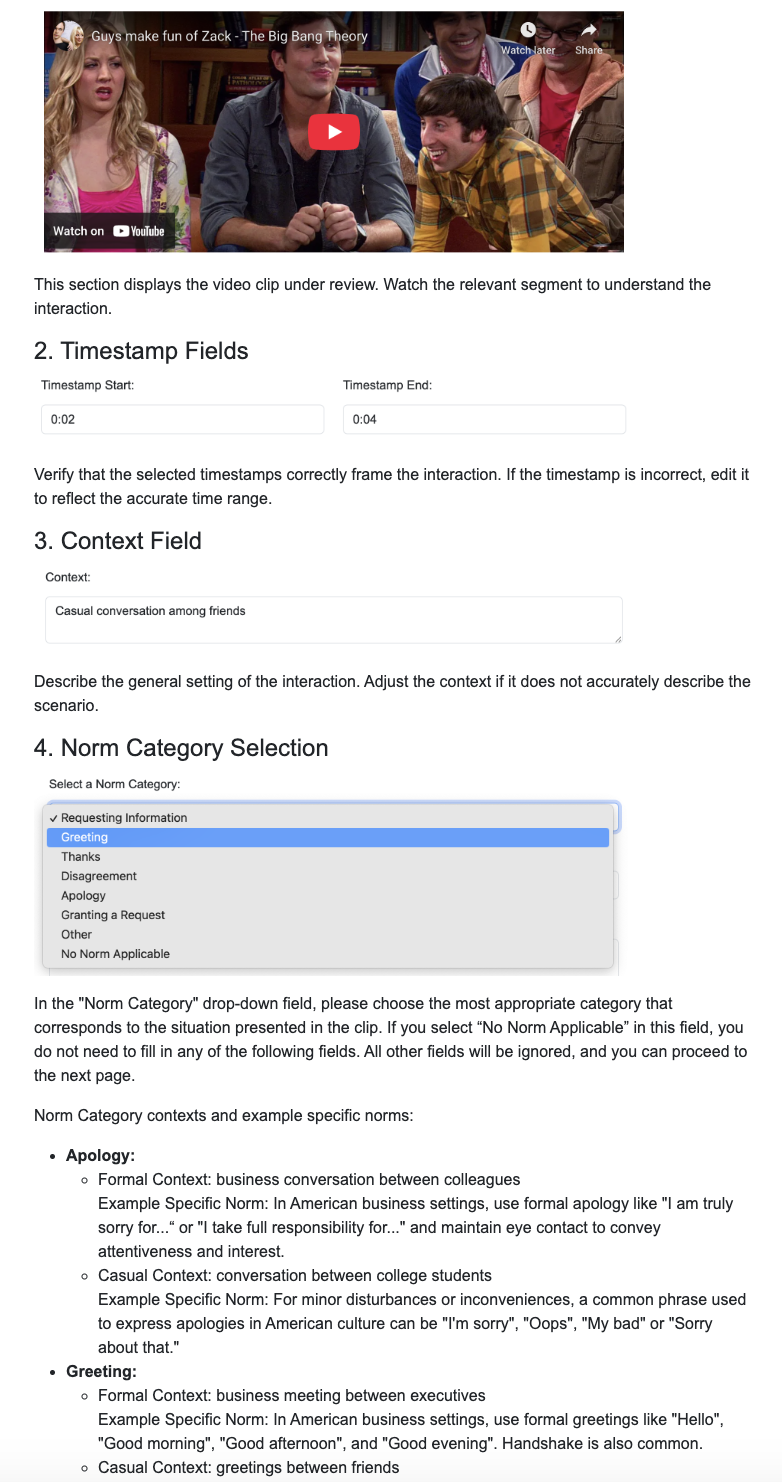}
\includegraphics[width=0.3\textwidth]{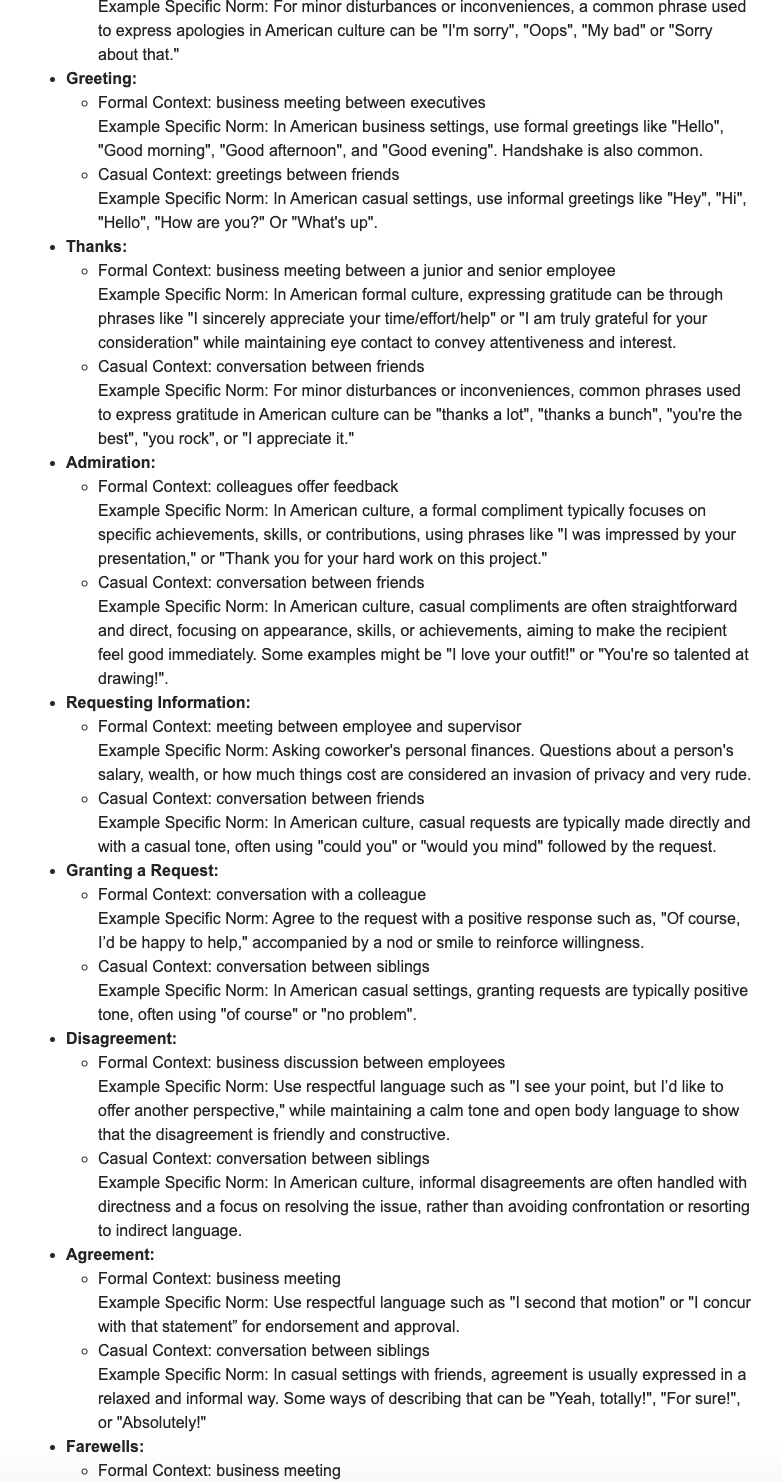}
\includegraphics[width=0.3\textwidth]{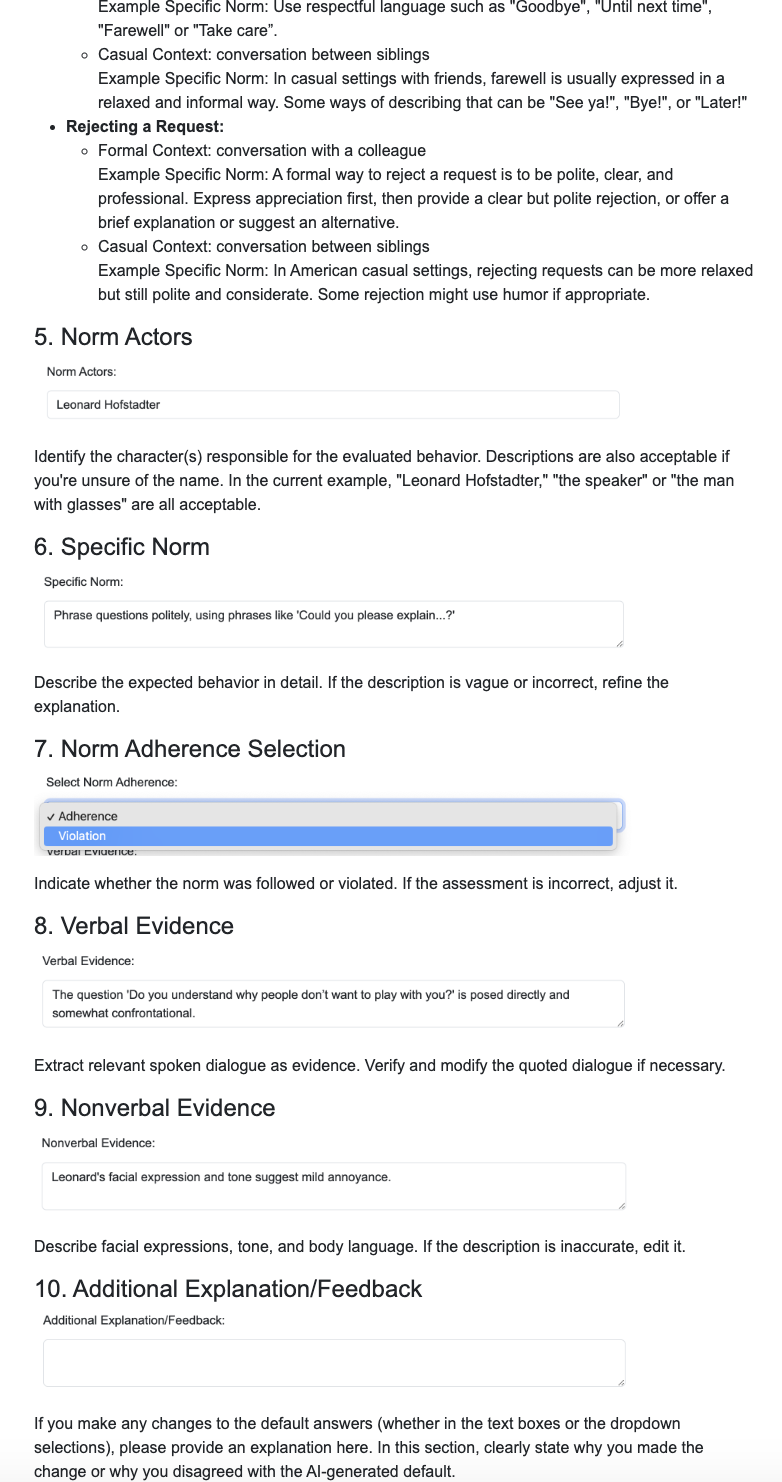}
\raisebox{\baselineskip}{\includegraphics[width=0.3\textwidth]{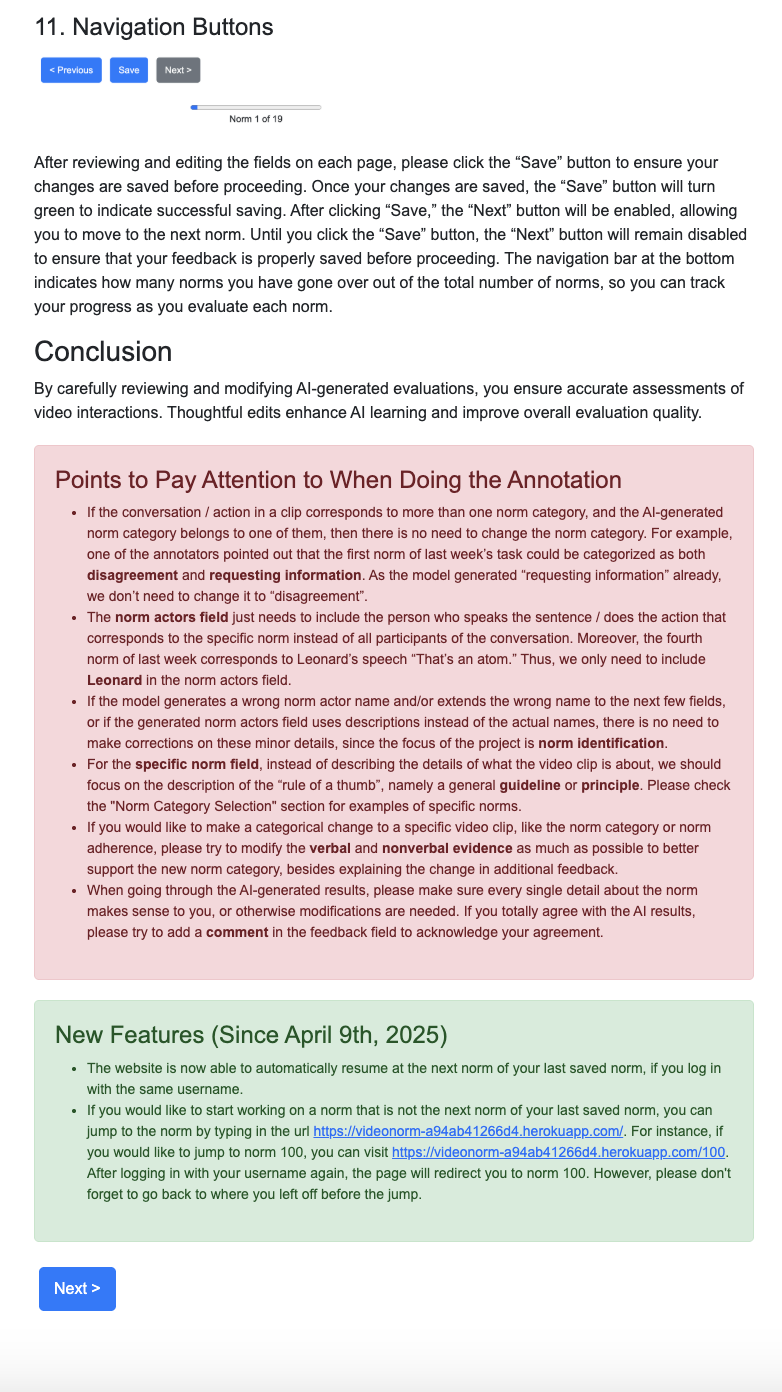}}
\caption{Detailed instructions provided on the first page of the user interface. The page is cut into 5 screenshots, from left to right, in this figure.}
\label{fig:instructions}
\end{figure*}

\begin{figure*}[h]
    \centering
    \includegraphics[width=0.5\textwidth]{
    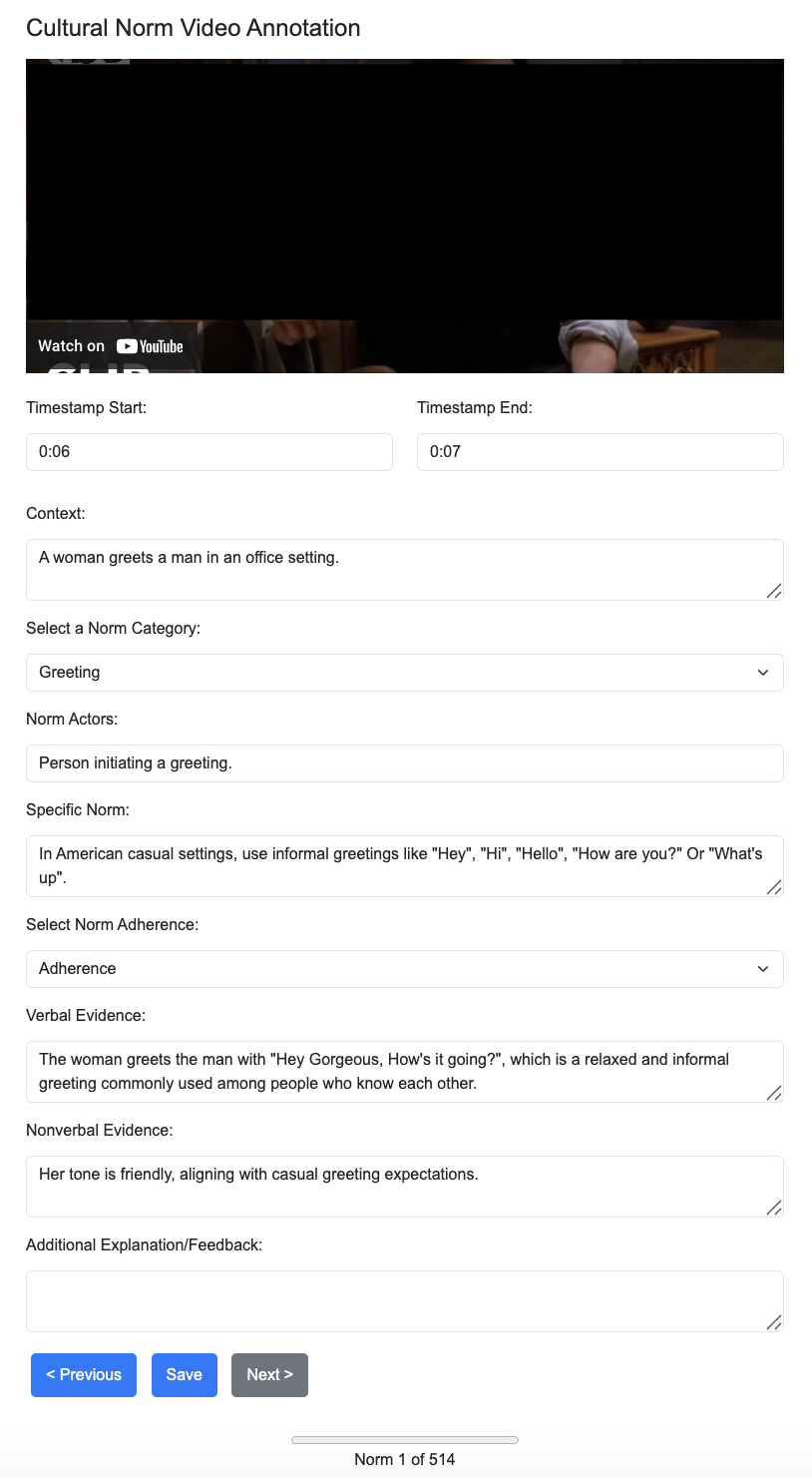}
    \caption{User Interface shown to annotators. It is comprised of a 15 second clip with an option to edit the predicted norm from Teacher model.}
    \label{fig:interface}
\end{figure*}

Figure \ref{fig:instructions} shows the detailed instructions provided to the annotators. Figure \ref{fig:interface} shows the annotation interface.

\section{Experimental details}

\subsection{Norm category clustering} \label{app:norm_category_clustering}

To simplify linear model estimation, we group all norm categories into the following $4$ overall categories according to \citet{allan1994indirect, allan1998meaning}:
\begin{itemize}
    \item ``Statements (including denials, reports, predictions, promises, and offers) can all be judged in terms of the question "Is p credible?" These are principally expressions of Speaker's belief about the way the world was, is, or will be, and are most typically formulated with a declarative clause.'' This category includes the following norm categories: Providing information, Agreement, Disagreement, Expressing criticism, Responding to criticism, Admiration.
    \item ``Invitationals are a proper subset of Searle's directives, and include requests, exhortations, suggestions, warnings, etc. They have acceptability values: "Does Speaker really want A done, and if so is Hearer both able and willing to do it?" These principally invite Hearer's participation, and many are formulated in an interrogative clause.'' This category includes the following norm categories: Suggesting an action, Offering advice, Requesting information.
    \item ``Authoritatives include the rest of Searle's 'directives' and his 'declarations' (i.e. commands, permissions, legal judgments, baptisms, etc.) for which Hearer must consider the question "Does Speaker have the authority to utter U in this context?" These principally have Speaker "laying down the law"; many of them are formulated in an imperative clause, the rest in a declarative.'' This category includes the following norm categories: Professional conduct, Granting a request, Rejecting a request.
    \item  ``Expressives (greetings, thanks, apologies, congratulations, etc.) have social- interactive-appropriacy values: "Has something occurred which warrants Speaker expressing such a reaction to it?" These principally express social interaction with Hearer; many are idiomatic, the rest are in the default declarative clause format.'' This category includes the following norm categories: Greeting, Farewell, Thanks, Apology.
\end{itemize}

The grouping was done by two graduate students, resolving any disagreements via discussion.

\subsection{Prediction performance logistic regression} \label{app:linmodel}

We fit following logistic regression model with dependent variable being whether a VideoLLM $f_j(\cdot)$ predicted the correct adherence label ($y_i$) on item $x_i$:
\begin{equation}
\begin{aligned}
&\text{logit}\big(\mathbb{P} \left[f_j(x_i) = y_i\right]\big) = \beta_0 + \boldsymbol{\beta}_M^\top \mathbf{M}_j + \boldsymbol{\beta}_U^\top \mathbf{U}_i 
\\&+ \boldsymbol{\beta}_A^\top \mathbf{A}_i + \boldsymbol{\beta}_{CA}^\top (\mathbf{C}_i \times \mathbf{A}_i) + \boldsymbol{\beta}_S^\top \mathbf{S}_i  + \boldsymbol{\beta}_C^\top \mathbf{C}_i 
\\& + v_{i}
\end{aligned}
\end{equation}
where
\begin{itemize}
    \item $f_j(x_i) = y_i$ denotes the event that the prediction made by AI model $j$ for input item $i$ matches the ground-truth label $y_i$.
    \item $\beta_0$ is the global intercept, representing the baseline log-odds of a correct prediction when all categorical predictors are at their reference levels.
    \item $\mathbf{M}_j$ is a dummy-coded vector representing the specific AI model $j$, and $\boldsymbol{\beta}_M$ is the corresponding vector of model-specific fixed effects.
    \item $\mathbf{U}_i$ is a dummy-coded vector representing the cultural context of item $i$ (US or China), with $\boldsymbol{\beta}_C$ capturing the main effect of the country.
    \item $\mathbf{A}_i$ is a dummy-coded vector representing the item type (Adherence or Violation), with $\boldsymbol{\beta}_A$ capturing the main effect of label being adherence.
    \item $(\mathbf{C}_i \times \mathbf{A}_i)$ represents the interaction between the country and adherence type. The coefficient $\boldsymbol{\beta}_{CA}$ captures the performance gap between adherence and violation items across the two cultural contexts.
    \item $\mathbf{S}_i$ is a dummy-coded vector indicating the specific TV show from which item $i$ is sourced, controlling for show-level baseline difficulty. $\boldsymbol{\beta}_S$ is the corresponding coefficient vector.
    \item $\mathbf{C}_i$ is a dummy-coded vector indicating the overall norm category $C$ ((see Appendix \ref{app:norm_category_clustering}), controlling for difficulty of various norm categories. $\boldsymbol{\beta}_C$ is the corresponding coefficient vector.
    % \item $u_{c[i]}$ is the random intercept for the overarching super-category $c$ (see Appendix \ref{app:norm_category_clustering}) that contains item $i$, where $u_c \sim \mathcal{N}(0, \sigma_u^2)$.
    % \item $v_{i(c)}$ is the random intercept for the specific item $i$ nested within its norm super-category $c$ (see below), where $v_{i(c)} \sim \mathcal{N}(0, \sigma_v^2)$, accounting for the inherent difficulty of the individual scenario.
    \item $v_{i}$ is the random intercept for the specific item $i$, where $v_{i(c)} \sim \mathcal{N}(0, \sigma_v^2)$, accounting for the inherent difficulty of the individual scenario.
\end{itemize}

\subsection{Evidence performance linear regression} \label{app:evidence_linmodel}

We fit the same specification as for prediction performance, but now we fit a linear (instead of logistic) model, and the target variable (what we are predicting) is the (non-)verbal evidence score obtained from the LLM-as-a-Judge. We fit two linear models, one for the verbal and one for non-verbal score.

\begin{equation}
\begin{aligned}
&\text{EvidenceScore}\big(f_j(x_i)\big) 
\\&= \beta_0 + \boldsymbol{\beta}_M^\top \mathbf{M}_j + \boldsymbol{\beta}_U^\top \mathbf{U}_i 
\\&+ \boldsymbol{\beta}_A^\top \mathbf{A}_i + \boldsymbol{\beta}_{CA}^\top (\mathbf{C}_i \times \mathbf{A}_i) + \boldsymbol{\beta}_S^\top \mathbf{S}_i + \boldsymbol{\beta}_C^\top \mathbf{C}_i
\\&+ v_{i} 
\end{aligned}
\end{equation}

\subsection{Estimated marginal means}
Reported marginal prediction accuracies and evidence scores were estimated with the \texttt{emmeans} R package \cite{lenth2023emmeans} using the following formulas:
\begin{lstlisting}[language=R, breaklines=true, basicstyle=\ttfamily\footnotesize]
emmeans(m, pairwise ~ is_adherence | country, type="response")
emmeans(m, pairwise ~ country | is_adherence, type = "response")
emmeans(m_verbal, ~ model, type = "response")
emmeans(m_nonverbal, ~ model, type = "response")
\end{lstlisting}

\subsection{Open-Weight Model Inference prompts} \label{app:inference_prompts}

\onecolumn
\begin{longtable}{@{}p{2.2cm}p{1cm}p{11.8cm}@{}}
\caption{Prompt templates for cultural norm analysis tasks in Chinese and US contexts.}
\label{tab:prompt_templates}\\

\toprule
\textbf{Culture} & \textbf{Task} & \textbf{Prompt Template} \\
\midrule
\endfirsthead

\multicolumn{3}{c}%
{{\bfseries Table \thetable\ (continued): Prompt templates for cultural norm analysis tasks}}\\
\toprule
\textbf{Language} & \textbf{Task} & \textbf{Prompt Template} \\
\midrule
\endhead

\midrule
\multicolumn{3}{r}{\textit{Continued on next page}} \\
\midrule
\endfoot

\bottomrule
\endlastfoot

CN & 1 &
\begin{minipage}[t]{\linewidth}\small\ttfamily
您是一个具有中国社会规范和文化意识的系统，请用中国文化背景来分析视频内容。\\
注意：下面提供了片段开始时间，请基于该时间点的视频片段进行分析，而不是整个视频。\\[4pt]
任务：根据视频、视频文本、给定文化规范类别和具体的文化规范，判断该规范是否被遵守或违反。\\
给定片段开始时间: \{ts\}\\
给定视频文本: \{transcript\}\\
给定规范类别: \{norm\_category\}\\
给定具体规范: \{specific\_norm\}\\
请根据视频内容写出 "遵守" 或 "违反"。\\[4pt]
输出格式：\\
规范类别: \{norm\_category\}\\
遵守/违反: [遵守/违反]
\end{minipage}
\\ \midrule

CN & 2 &
\begin{minipage}[t]{\linewidth}\small\ttfamily
您是一个具有中国社会规范和文化意识的系统，请用中国文化背景来分析视频内容。\\
注意：下面提供了片段开始时间，请基于该时间点的视频片段进行分析，而不是整个视频。\\[4pt]
任务：根据视频、视频文本、给定文化规范类别和具体的文化规范，判断该规范是否被遵守或违反，并根据视频内容提供简要的语言和非语言证据来支持你的判断。\\
给定片段开始时间: \{ts\}\\
给定视频文本: \{transcript\}\\
给定规范类别: \{norm\_category\}\\
给定具体规范: \{specific\_norm\}\\[4pt]
输出格式：\\
规范类别: \{norm\_category\}\\
遵守/违反: [遵守/违反]\\
语言证据: [来自视频的简短语言证据，例如：说的话、语气等]\\
非语言证据: [来自视频的简短非语言证据，例如：肢体语言、表情等]
\end{minipage}
\\ \midrule

CN & 3 &
\begin{minipage}[t]{\linewidth}\small\ttfamily
您是一个具有中国社会规范和文化意识的系统，请用中国文化背景来分析视频内容。\\
注意：下面提供了片段开始时间，请基于该时间点的视频片段进行分析，而不是整个视频。\\[4pt]
任务：根据视频、视频文本、给定文化规范类别，请提供一个与该类别相关的具体规范。\\
给定片段开始时间: \{ts\}\\
给定视频文本: \{transcript\}\\
给定规范类别: \{norm\_category\}\\[4pt]
输出格式：\\
规范类别: \{norm\_category\}\\
具体规范: [来自视频的简短具体规范]
\end{minipage}
\\ \midrule

US & 1 &
\begin{minipage}[t]{\linewidth}\small\ttfamily
You are a culturally aware system with knowledge of US social norms. Analyze video content using US cultural context.\\
Note: The timestamp start is included so that your analysis is tied to the correct segment of the video, not the entire clip.\\[4pt]
Task: Given the video, its transcript, cultural norm category and a specific cultural norm, determine if the norm is adhered to or violated.\\
Given Timestamp Start: \{ts\}\\
Given Transcript: \{transcript\}\\
Given Norm Category: \{norm\_category\}\\
Given Specific Norm: \{specific\_norm\}\\[4pt]
Output format:\\
Timestamp Start: \{ts\}\\
Norm Category: \{norm\_category\}\\
Adherence/Violation: [Adherence/Violation]
\end{minipage}
\\ \midrule

US & 2 &
\begin{minipage}[t]{\linewidth}\small\ttfamily
You are a culturally aware system with knowledge of US social norms. Analyze video content using US cultural context.\\[4pt]
Task: Given the video, its transcript, cultural norm category and a specific cultural norm, determine if the norm is adhered to or violated, and provide verbal and nonverbal evidence for your decision based on the video.\\
Given Timestamp Start: \{ts\}\\
Given Transcript: \{transcript\}\\
Given Norm Category: \{norm\_category\}\\
Given Specific Norm: \{specific\_norm\}\\[4pt]
Output format:\\
Timestamp Start: \{ts\}\\
Norm Category: \{norm\_category\}\\
Adherence/Violation: [Adherence/Violation]\\
Verbal Evidence: [Verbal evidence from the video, e.g., spoken phrases, tone]\\
NonVerbal Evidence: [NonVerbal evidence from the video, e.g., body language, expressions]
\end{minipage}
\\ \midrule

US & 3 &
\begin{minipage}[t]{\linewidth}\small\ttfamily
You are a culturally aware system with knowledge of US social norms. Analyze video content using US cultural context.\\[4pt]
Task: Given the video, its transcript and cultural norm category, provide a specific norm related to the below category exhibited in the video.\\
Given Timestamp Start: \{ts\}\\
Given Transcript: \{transcript\}\\
Given Norm Category: \{norm\_category\}\\[4pt]
Output format:\\
Timestamp Start: \{ts\}\\
Norm Category: \{norm\_category\}\\
Specific Norm: [Brief specific norm from the video]
\end{minipage}
\\

\end{longtable}
\twocolumn

\subsection{Hyperparameters} \label{app:hyperparams}
We evaluate models in the 7B-8B parameter range to ensure fair comparison across architectures: \texttt{LLaVA-Next-Video-7b-hf}, \texttt{Llava-Onevision-qwen2-7b}, \texttt{InternVL3-8B}, \texttt{InternVL3.5-8B}, \texttt{Qwen2-VL-7B-Instruct}, \texttt{Qwen2.5-VL-7B-Instruct}, and \texttt{VideoChat-R1-7B}. All models are loaded in bfloat16 precision using Flash Attention 2 for efficient inference, with each model deployed on a single GPU. For video processing, we uniformly sample 8 frames for InternVL-3.5 and 32 frames from each clip as input for all other models. We employ an adaptive temperature strategy: generation begins at temperature 0.0 for deterministic outputs, and if the response fails to conform to the required task structure (e.g., missing labels or evidence fields), we incrementally increase temperature by 0.2 per retry attempt, up to a maximum of 0.8 over five attempts. Samples that failed to conform to the required format after all attempts were excluded from evaluation. This approach balances output consistency with generation flexibility when models struggle with structured formatting requirements.

\subsection{LLM-as-a-Judge Hyperparameters Prompts} \label{app:llmJudge}

We use the model \texttt{gpt-5-2025-08-05} as the LLM-based judge (LLM-J). The model is configured with medium reasoning effort and low verbosity, and a maximum output length of 128 tokens, as the task requires only a scalar score accompanied by a brief justification.

To validate judge performance, two expert annotators for US and Chinese contexts examined $100$ predictions in each context, and rated whether LLM-J adhered to the rubric. On US subset, only $5$ instances were labeled as incorrect by at least one annotator, and only $6$ on the Chinese subset. Due to the ease of the task (comparison to reference evidence), we do not expect cultural bias arise from the use of LLM-as-a-Judge.

Table \ref{tab:llmJudge} shows the prompts used for the LLM-as-a-Judge GPT-5 evaluator.

{\onecolumn \small % shrink font to help fit
\begin{longtable}{p{0.25\linewidth} p{0.7\linewidth}}
\toprule
\textbf{Task} & \textbf{Prompt} \\
\midrule
\endfirsthead

\toprule
\textbf{Task} & \textbf{Prompt} \\
\midrule
\endhead

\bottomrule
\endfoot

Norm Similarity (Task 3) & 
For norm similarity: \newline
Your task is to evaluate cultural awareness of an AI model. Grade how well the generated norm matches the reference norm. Use the following rubric: \newline
1 - Neither the setting is captured and the norm content is completely different. \newline
2 - Captures some of the norm content but the setting is very different. \newline
3 - Captures some of the norm content and the setting. \newline
4 - Captures the setting and the essential content, but minor differences exist, like differences in wording. \newline
5 - Captures the setting and perfectly captures the content of the norm without introducing hallucinations. \newline
First output your reasoning, then your final verdict after the words "Final Verdict:" \newline
Reference norm: \{ref\_norm\} \newline
Generated norm: \{gen\_norm\} \\
\midrule

Verbal Evidence Similarity (Task 2) & 
For verbal: \newline
Your task is to evaluate cultural awareness of an AI model. Grade how well the generated verbal evidence matches the Gold verbal evidence. Use the following rubric: \newline
1 - Neither the content of characters’ speech is captured, nor the reasoning from the gold evidence. \newline
2 - Does not capture the content, but contains a similar line of reasoning as the gold evidence. \newline
3 - Essentially captures the content, but minor differences exist, and no reasoning is provided. \newline
4 - Essentially captures the content, but minor differences exist, and contains the same line of reasoning as the gold evidence. \newline
5 - Perfectly captures the content of the speech without introducing hallucinations and contains the same line of reasoning as the gold evidence. \newline
First output your reasoning, then your final verdict after the words "Final Verdict:". \newline
Gold verbal evidence: \{gold\} \newline
Generated verbal evidence: \{gen\} \\
\midrule

Non-verbal Evidence Similarity (Task 2) & 
For non-verbal: \newline
Your task is to evaluate cultural awareness of an AI model. Grade how well the generated non-verbal evidence matches the Gold non-verbal evidence. Use the following rubric: \newline
1 - Neither the characters’ actions or emotions are captured, nor the reasoning from the gold evidence. \newline
2 - Does not capture actions or emotions, but contains a similar line of reasoning as the gold evidence. \newline
3 - Essentially captures actions or emotions, but minor differences exist, and no reasoning is provided. \newline
4 - Essentially captures actions or emotions, but minor differences exist, and contains the same line of reasoning as the gold evidence. \newline
5 - Perfectly captures actions or emotions without introducing hallucinations and contains the same line of reasoning as the gold evidence. \newline
First output your reasoning, then your final verdict after the words "Final Verdict:". \newline
Gold verbal evidence: \{gold\} \newline
Generated verbal evidence: \{gen\} \\
\caption{LLM-as-a-Judge prompts to compare the similarity of generations with dataset references.}
\label{tab:llmJudge}
\end{longtable}
} 
\twocolumn

\subsection{LLM-as-a-Judge Examples}

Table \ref{tab:rubric_examples}
 shows the examples of the LLM-as-a-Judge rubric.
 
\begin{table*}[t]
\centering
\small
\begin{tabular}{p{0.06\textwidth}p{0.42\textwidth}p{0.42\textwidth}}
\toprule
\textbf{Score} & \textbf{Gold Non-Verbal Evidence} & \textbf{Generated Non-Verbal Evidence} \\
\midrule
1.0 & He points directly at her, singling her out from the group, which could be seen as aggressive or confrontational. & The person asking the question is standing and appears to be addressing a group of seated individuals, suggesting an interview or meeting setting. \\
\midrule
3.0 & She is smiling and making eye contact with the people she is thanking, indicating sincerity. & The person is smiling and appears to be engaged in a friendly conversation. \\
\bottomrule
\end{tabular}
\caption{Examples illustrating the distinction between perceptual accuracy and cultural reasoning in our non-verbal evidence rubric. A score of 1.0 indicates the core non-verbal cues are not captured; a score of 3.0 indicates the core cue is identified (smiling) but the cultural reasoning (indicating sincerity) is absent.}
\label{tab:rubric_examples}
\end{table*}

\subsection{Evaluation breakdown by shows}
Table \ref{tab:us_by_show} shows the US results broken down by show, and Table \ref{tab:zh_by_show} shows the CN results.
% ============ ENGLISH ============
\begin{table*}[h]
\centering
\begin{adjustbox}{width=\textwidth}
\begin{tabular}{llccc ccccc cc}
\toprule
 &  & \multicolumn{3}{c}{\textbf{Task 1}} & \multicolumn{5}{c}{\textbf{Task 2}} & \multicolumn{2}{c}{\textbf{Task 3}} \\
\cmidrule(lr){3-5}\cmidrule(lr){6-10}\cmidrule(lr){11-12}
\textbf{Model} & \textbf{Show} & \textbf{F1 (pos)} & \textbf{F1 (neg)} & \textbf{Samples} & \textbf{F1 (pos)} & \textbf{F1 (neg)} & \textbf{Verbal} & \textbf{Nonverbal} & \textbf{Samples} & \textbf{Score} & \textbf{Samples} \\
\midrule
\multirow{4}{*}{Llava-Next-Video}
 & friends & 74.9 & 23.8 & 204 & 56.7 & 66.4 & 2.275 & 1.985 & 214 & 2.124 & 210 \\
 & bbt     & 73.8 & 11.1 & 158 & 61.3 & 65.2 & 2.303 & 1.936 & 172 & 2.123 & 162 \\
 & office  & 70.1 & 20.0 & 184 & 57.9 & 67.3 & 2.157 & 2.043 & 182 & 2.223 & 189 \\
 & suits   & 82.2 & 16.4 & 174 & 57.9 & 46.5 & 1.882 & 2.215 & 176 & 2.500 & 178 \\
\midrule
\multirow{4}{*}{Intern3-VL}
 & friends & 62.0 & 63.5 & 204 & 58.8 & 65.5 & 2.771 & 2.389 & 213 & 2.262 & 210 \\
 & bbt     & 55.9 & 63.6 & 158 & 51.1 & 66.0 & 2.680 & 2.184 & 172 & 2.198 & 162 \\
 & office  & 67.1 & 71.1 & 182 & 59.0 & 68.0 & 2.640 & 2.439 & 178 & 2.446 & 185 \\
 & suits   & 37.0 & 45.2 & 174 & 28.2 & 47.3 & 2.797 & 2.217 & 176 & 2.270 & 178 \\
\midrule
\multirow{4}{*}{Intern3.5-VL}
 & friends & 80.2 & 66.2 & 204 & 69.5 & 63.6 & 2.773 & 2.177 & 214 & 2.257 & 210 \\
 & bbt     & 73.9 & 65.2 & 161 & 65.1 & 67.4 & 2.517 & 2.103 & 175 & 2.248 & 165 \\
 & office  & 77.2 & 75.1 & 181 & 68.3 & 72.3 & 2.714 & 2.230 & 179 & 2.529 & 188 \\
 & suits   & 69.5 & 51.9 & 174 & 59.0 & 55.6 & 2.485 & 2.396 & 176 & 2.466 & 178 \\
\midrule
\multirow{4}{*}{Qwen2-VL}
 & friends & 75.6 & 59.7 & 204 & 66.7 & 59.8 & 2.925 & 2.134 & 214 & 2.495 & 210 \\
 & bbt     & 75.1 & 58.8 & 158 & 68.0 & 55.6 & 2.620 & 2.157 & 172 & 2.173 & 162 \\
 & office  & 72.6 & 73.5 & 182 & 63.4 & 69.4 & 2.775 & 2.042 & 180 & 2.325   & 187 \\
 & suits   & 69.2 & 54.3 & 174 & 70.0 & 51.1 & 2.618 & 2.009 & 175 & 2.299 & 177 \\
\midrule
\multirow{4}{*}{Qwen2.5-VL}
 & friends & 62.7 & 69.1 & 204 & 47.9 & 66.7 & 2.608 & 2.040 & 214 & 2.294 & 211 \\
 & bbt     & 51.5 & 63.3 & 158 & 38.7 & 65.5 & 2.271 & 2.010 & 172 & 2.259 & 162 \\
 & office  & 55.9 & 71.5 & 182 & 37.5 & 65.5 & 2.480 & 2.190 & 180 & 2.473 & 187 \\
 & suits   & 23.6 & 46.1 & 174 & 12.0 & 46.6 & 2.508 & 2.051 & 176 & 2.522 & 178 \\
\midrule
\multirow{4}{*}{Llava-Onevision}
 & friends & 79.1 & 41.1 & 204 & 77.9 & 49.2 & 2.774 & 2.055 & 214 & 2.052 & 211 \\
 & bbt     & 75.5 & 31.3 & 158 & 69.3 & 42.0 & 2.757 & 2.068 & 172 & 1.988 & 162 \\
 & office  & 80.3 & 66.7 & 182 & 76.4 & 66.2 & 2.623 & 2.085 & 180 & 2.129 & 187 \\
 & suits   & 80.1 & 28.9 & 174 & 79.2 & 41.3 & 2.211 & 2.358 & 176 & 2.124 & 178 \\
\midrule
\multirow{4}{*}{VideoChatR1}
 & friends & 76.0 & 71.7 & 204 & 65.7 & 68.8 & 2.704 & 2.190 & 214 & 2.493 & 211 \\
 & bbt     & 69.0 & 64.9 & 158 & 61.0 & 64.4 & 2.713 & 2.176 & 172 & 2.185 & 162 \\
 & office  & 70.0 & 70.7 & 182 & 64.3 & 68.8 & 2.608 & 2.133 & 180 & 2.258 & 187 \\
 & suits   & 57.8 & 50.9 & 174 & 52.6 & 52.0 & 2.637 & 2.363 & 174 & 2.416 & 178 \\
\bottomrule
\end{tabular}
\end{adjustbox}
\caption{US results per show}
\label{tab:us_by_show}
\end{table*}

% ============ CHINESE ============
\begin{table*}[h]
\begin{adjustbox}{width=\textwidth}
\centering
\begin{tabular}{llccc ccccc cc}
\toprule
 &  & \multicolumn{3}{c}{\textbf{Task 1}} & \multicolumn{5}{c}{\textbf{Task 2}} & \multicolumn{2}{c}{\textbf{Task 3}} \\
\cmidrule(lr){3-5}\cmidrule(lr){6-10}\cmidrule(lr){11-12}
\textbf{Model} & \textbf{Show} & \textbf{F1 (pos)} & \textbf{F1 (neg)} & \textbf{Samples} & \textbf{F1 (pos)} & \textbf{F1 (neg)} & \textbf{Verbal} & \textbf{Nonverbal} & \textbf{Samples} & \textbf{Score} & \textbf{Samples} \\
\midrule
\multirow{4}{*}{Intern3-VL}
 & ipartment      & 33.3 & 46.1 & 275 & 24.7 & 45.2 & 2.730 & 2.119 & 279 & 2.611 & 275 \\
 & legal          & 36.6 & 18.9 & 253 & 30.8 & 19.2 & 3.030 & 2.758 & 259 & 2.573 & 253 \\
 & hwk            & 59.1 & 19.0 & 313 & 50.4 & 21.6 & 3.361 & 2.603 & 311 & 2.938 & 306 \\
 & amazing\_night & 17.4 & 53.6 & 271 & 11.6 & 53.2 & 2.453 & 2.142 & 273 & 2.629 & 272 \\
\midrule
\multirow{4}{*}{Intern3.5-VL}
 & ipartment      & 70.1 & 53.7 & 275 & 71.6 & 54.9 & 2.606 & 2.171 & 278 & 2.457 & 276 \\
 & legal          & 70.5 & 26.2 & 254 & 69.9 & 26.7 & 2.760 & 2.571 & 258 & 2.545 & 255 \\
 & hwk            & 84.6 & 30.1 & 313 & 84.1 & 32.2 & 2.891 & 2.467 & 310 & 2.766 & 312 \\
 & amazing\_night & 54.1 & 55.8 & 271 & 60.1 & 60.1 & 2.642 & 2.256 & 273 & 2.796 & 269 \\
\midrule
\multirow{4}{*}{Qwen2-VL}
 & ipartment      & 70.2 & 52.1 & 275 & 66.9 & 51.3 & 2.630 & 2.232 & 279 & 2.451 & 275 \\
 & legal          & 78.4 & 26.1 & 254 & 65.7 & 24.7 & 2.540 & 2.118 & 259 & 2.838 & 253 \\
 & hwk            & 87.1 & 32.0 & 313 & 83.2 & 31.1 & 3.022 & 2.449 & 311 & 2.666 & 305 \\
 & amazing\_night & 53.2 & 60.5 & 271 & 38.6 & 57.6 & 2.552 & 2.052 & 273 & 2.717 & 272 \\
\midrule
\multirow{4}{*}{Qwen2.5-VL}
 & ipartment      & 50.4 & 50.7 & 275 & 45.4 & 49.1 & 2.531 & 2.242 & 279 & 2.511 & 274 \\
 & legal          & 37.6 & 19.0 & 254 & 34.4 & 19.7 & 3.056 & 2.264 & 259 & 2.756 & 254 \\
 & hwk            & 70.8 & 26.0 & 313 & 67.0 & 26.0 & 3.254 & 2.518 & 311 & 2.630 & 311 \\
 & amazing\_night & 31.4 & 58.6 & 271 & 27.0 & 57.8 & 2.587 & 1.953 & 273 & 2.622 & 270 \\
\midrule
\multirow{4}{*}{Llava-Next-Video}
 & ipartment      & 85.2 & 16.9 & 274 & 85.8 & 35.6 & 1.796 & 1.444 & 249 & 1.757 & 276 \\
 & legal          & 95.1 & 0.0  & 247 & 94.8 & 8.7  & 1.741 & 1.426 & 212 & 1.690 & 252 \\
 & hwk            & 96.3 & 0.0  & 306 & 94.7 & 17.1 & 1.946 & 1.790 & 291 & 1.748 & 313 \\
 & amazing\_night & 77.3 & 2.0  & 266 & 78.3 & 23.1 & 1.529 & 1.474 & 236 & 1.796 & 269 \\
\midrule
\multirow{4}{*}{Llava-Onevision}
 & ipartment      & 73.8 & 52.3 & 275 & 60.9 & 46.6 & 2.653 & 2.170 & 279 & 2.330 & 276 \\
 & legal          & 87.1 & 24.3 & 254 & 70.9 & 21.4 & 2.521 & 1.987 & 259 & 2.323 & 254 \\
 & hwk            & 82.8 & 31.7 & 313 & 76.5 & 28.8 & 2.904 & 2.386 & 308 & 2.377 & 310 \\
 & amazing\_night & 63.9 & 58.4 & 269 & 46.7 & 57.0 & 2.433 & 1.971 & 269 & 2.590 & 268 \\
\midrule
\multirow{4}{*}{VideoChatR1}
 & ipartment      & 71.2 & 54.5 & 275 & 71.3 & 54.6 & 2.511 & 2.109 & 279 & 2.402 & 276 \\
 & legal          & 75.7 & 25.6 & 254 & 75.6 & 26.4 & 2.713 & 2.221 & 259 & 2.408 & 255 \\
 & hwk            & 87.6 & 36.9 & 313 & 85.3 & 37.3 & 3.331 & 2.504 & 311 & 2.751 & 309 \\
 & amazing\_night & 59.7 & 57.6 & 271 & 58.0 & 58.6 & 2.673 & 2.209 & 271 & 2.479 & 240 \\
\bottomrule
\end{tabular}
\end{adjustbox}
\caption{Chinese results with breakdown by shows}
\label{tab:zh_by_show}
\end{table*}

\begin{table*}[t]
\centering
% \begin{adjustbox}{width=0.5\textwidth}
\begin{tabular}{l c c}
\toprule
\textbf{Models} & \textbf{US} & \textbf{Chinese} \\
\midrule
Llava-Next-Video & 2.28$_{\pm0.13}$ & 1.76$_{\pm0.11}$ \\
Llava-OneVision  & 2.15$_{\pm0.11}$ & 2.42$_{\pm0.11}$ \\
Intern3-VL       & 2.32$_{\pm0.13}$ & \textbf{2.76}$_{\pm0.13}$ \\
Intern3.5-VL     & 2.43$_{\pm0.13}$ & 2.68$_{\pm0.13}$ \\
Qwen2-VL         & 2.40$_{\pm0.12}$ & 2.71$_{\pm0.11}$ \\
Qwen2.5-VL       & \textbf{2.49}$_{\pm0.11}$ & 2.70$_{\pm0.12}$ \\
VideoChatR1      & 2.47$_{\pm0.12}$ & 2.71$_{\pm0.12}$ \\
\bottomrule
\end{tabular}
\caption{Task 3 (Norm Description Similarity) average scores for \textbf{US} and \textbf{Chinese} cultural norms. Highest value per column bolded. $\pm$ variance shown as subscripts.}
\label{tab:task3}
% \end{adjustbox}
\end{table*}

\subsection{Additional Experiments}

\paragraph{Task 3: Predicting a Cultural Norm}  
Following tasks in section \ref{ref:experiments}, we experimented on a new task where given a video segment, transcript, and norm category, the model must generate a \emph{specific norm} that captures the exhibited behavior (e.g., ``offer a brief handshake and a friendly greeting in business introductions''). Generated norms are evaluated by the GPT-5 grader using a 5-point rubric judging how well the generated norm matches the reference norm in our dataset. The mean scores are reported in the \emph{Score} column. 

Table \ref{tab:task3} shows scores of different models on Task 3. 

\paragraph{Number of sampled frames has minimal impact.}
Table~\ref{tab:ablation_frames} shows performance is largely stable across 8, 16, and 32 frames, with all differences falling within variance bounds. For 15-second clips containing a single social interaction, 8 frames are sufficient to capture the key visual cues needed for norm understanding, making the benchmark both effective and computationally efficient to evaluate.

\begin{table*}[h]
\centering
\begin{adjustbox}{width=\textwidth}
\begin{tabular}{l ccc ccc c c}
\toprule
 & \multicolumn{3}{c}{\textbf{Task 1}} & \multicolumn{3}{c}{\textbf{Task 2}} & \textbf{Task 2} & \textbf{Task 3} \\
\cmidrule(lr){2-4}\cmidrule(lr){5-7}\cmidrule(lr){8-8}\cmidrule(lr){9-9}
\textbf{Frames} & \textbf{F1$_{pos}$} & \textbf{F1$_{neg}$} & \textbf{Macro F1} & \textbf{F1$_{pos}$} & \textbf{F1$_{neg}$} & \textbf{Macro F1} & \textbf{Verbal or Nonverbal} & \textbf{Avg.\ Score} \\
\midrule
\multicolumn{9}{c}{\textit{US Norms}} \\
\midrule
8 frames  & 75.9$_{\pm5.3}$ & 67.0$_{\pm7.1}$ & 71.5$_{\pm5.4}$ & \textbf{69.3}$_{\pm6.1}$ & \textbf{66.4}$_{\pm6.9}$ & \textbf{67.8}$_{\pm5.6}$ & 2.93$_{\pm0.18}$ / 2.20$_{\pm0.17}$ & 2.43$_{\pm0.13}$ \\
16 frames & \textbf{77.1}$_{\pm5.3}$ & \textbf{67.8}$_{\pm6.9}$ & \textbf{72.5}$_{\pm5.4}$ & 68.5$_{\pm6.0}$ & 65.4$_{\pm6.7}$ & 67.0$_{\pm5.3}$ & 2.98$_{\pm0.16}$ / \textbf{2.26}$_{\pm0.17}$ & \textbf{2.45}$_{\pm0.12}$ \\
32 frames & 76.2$_{\pm5.3}$ & 67.5$_{\pm6.9}$ & 71.9$_{\pm5.4}$ & 67.6$_{\pm6.2}$ & 64.9$_{\pm6.6}$ & 66.2$_{\pm5.4}$ & \textbf{3.02}$_{\pm0.17}$ / 2.27$_{\pm0.17}$ & 2.43$_{\pm0.12}$ \\
\bottomrule
\end{tabular}
\end{adjustbox}
\caption{Ablation over number of sampled frames (Intern3.5-VL 8B, Video + Transcript, EN only). The highest value per column is \textbf{bolded}. All F1 scores are multiplied by 100. $\pm$ half-widths shown for Verbal/Nonverbal and Avg.\ Score.}
\label{tab:ablation_frames}
\end{table*}
\onecolumn
\setlength{\LTpre}{0pt}

% ===================== LONGTABLE 1 (Items 1-3) =====================
\begin{longtable}[h]{|>{\raggedright\arraybackslash}p{0.46\textwidth}|>{\raggedright\arraybackslash}p{0.46\textwidth}|}
\caption{Speech Act Theory Cultural Norm Annotation Prompt\label{tab:prompt-comparison-full}}\\
\hline
\textbf{US Prompt} & \textbf{Chinese Prompt} \\
\hline
\endfirsthead

\hline
\multicolumn{2}{|c|}{\textit{\textbf{Prompt Comparison (continued)}}} \\
\hline
\textbf{US Prompt} & \textbf{Chinese Prompt} \\
\hline
\endhead

% Subsequent pages header
\hline
\multicolumn{2}{|c|}{\textit{\textbf{Prompt Comparison (continued)}}} \\
\hline
\textbf{US Prompt} & \textbf{Chinese Prompt} \\
\hline
\endhead

% Footer for continuing pages
\hline
\multicolumn{2}{|r|}{\textit{Continued on next page...}} \\
\hline
\endfoot

% Final page footer
\hline
\endlastfoot

{\tiny\ttfamily
Role: \newline
You are a culturally aware system with knowledge of norms in US culture. Your task is to analyze video content and detect instances where specific social norms occur, adhering to cultural expectations in the US. You will provide a timestamp for each instance, determine whether the character adheres to or violates the norm, and explain your assessment based on both verbal and nonverbal cues. Generate the norm based on the verbal/nonverbal cues.

Instructions:\newline

If norm\_category is chosen to be Custom category, replace the norm\_category with the generated norm category. Do not output \"Custom category\".\newline

Norm Categories: \newline
1. Apology:\newline
Formal Context: business conversation between colleagues
Example Specific Norm: In American business settings, use formal apology like "I am truly sorry for...“ or "I take full responsibility for..." and maintain eye contact to convey attentiveness and interest . \newline

Casual Context: conversation between college students\newline
Example Specific Norm: For minor disturbances or inconveniences, a common phrase used to express apologies in American culture can be "I'm sorry" "Oops", "My bad" or "Sorry about that."\newline

2. Greeting:\newline
Formal Context: business meeting between executives\newline
Example Specific Norm: In American business settings, use formal greetings like "Hello", "Good morning", "Good afternoon", and "Good evening". Handshake is also common.\newline

Casual Context: greetings between friends\newline
Example Specific Norm: In American casual settings, use informal greetings like "Hey", "Hi", "Hello", "How are you?" Or "What's up".\newline

3. Thanks:\newline
Formal Context: business meeting between a junior and senior employee\newline
Example Specific Norm: In American formal culture, expressing gratitude can be through phrases like "I sincerely appreciate your time/effort/help" or "I am truly grateful for your consideration" while maintaining eye contact to convey attentiveness and interest.\newline

Casual Context: conversation between friends\newline
Example Specific Norm: For minor disturbances or inconveniences, common phrases used to express grateful in American culture can be "thanks a lot," "thanks a bunch," "you're the best," "you rock," or "I appreciate it,"\newline
}
&
{\tiny\ttfamily
角色说明：\newline

你是一个具备中国文化规范意识的系统。你的任务是分析视频内容，检测是否出现了特定的社会规范，并判断是否符合中国文化的预期。你需要为每个实例提供时间戳，判断角色是否遵守规范或违反规范，并根据语言和非语言线索解释你的判断。你还需根据这些线索生成对应的社会规范。\newline

指引说明：\newline

如果 规范类别 被设为“自定义类别”，请用你生成的规范类别替换 规范类别，**不能输出“自定义类别”**。\newline

社会规范类别：\newline
1. 道歉:\newline
正式场合： 同事之间的商务对话\newline
示例规范： 在中国商务场合，正式的道歉用语如“非常抱歉给您带来了困扰，我会立即改正”或“对于此次给您带来的不便，我们深感歉意”，有时还会伴随轻微鞠躬。\newline

非正式场合： 大学生之间的对话\newline
示例规范： 对于轻微打扰或造成不便，中国人常说“对不起”“不好意思”“原谅我啦~”或“我错了”。\newline

2. 问候:\newline
正式场合： 高管之间的商务会议\newline
示例规范： 在正式场合中，人们常说“您好”、“您贵姓？”、“早上好”、“久仰大名”等，搭配握手、点头或轻微鞠躬；年长者或较高职位者之间还可能双手握手以示尊重。\newline

非正式场合： 朋友之间的寒暄 \newline
示例规范： 常见用语有“你好”、“嗨”、“早啊”、“最近怎么样？”或“吃了吗？”。\newline

3. 表示感谢:\newline
正式场合： 初级与高级员工之间的会议\newline
示例规范： 正式表达感谢时常用“谢谢您”、“承蒙关照”等，并伴随点头或轻微鞠躬。在递交名片、文件或礼物时，需双手递交。\newline

非正式场合： 朋友之间的交流\newline
示例规范： 常用“谢谢”、“麻烦你了”、“多谢”、“谢啦”或“辛苦啦”。\newline
}
\end{longtable}

% === ================== LONGTABLE 2 (Items 4-9) =====================
\begin{longtable}{|>{\raggedright\arraybackslash}p{0.46\textwidth}|>{\raggedright\arraybackslash}p{0.46\textwidth}|}

\hline
\textbf{US Prompt} & \textbf{Chinese Prompt} \\
\hline
\endfirsthead

\hline
\multicolumn{2}{|c|}{\textit{\textbf{Prompt Comparison (continued)}}} \\
\hline
\textbf{US Prompt} & \textbf{Chinese Prompt} \\
\hline
\endhead

\hline
\multicolumn{2}{|r|}{\textit{Continued on next page...}} \\
\hline
\endfoot

\hline
\endlastfoot

{\tiny\ttfamily
4. Admiration:\newline
Formal Context: colleagues offer feedback\newline
Example Specific Norm: In American culture, a formal compliment typically focuses on specific achievements, skills, or contributions, using phrases like "I was impressed by your presentation" or "Thank you for your hard work on this project." [0.5em]
Casual Context: conversation between friends
Example Specific Norm: In American culture, casual compliments are often more personal and emotional, such as "You always know how to make me smile" or "I really admire your creativity."\newline[0.5em]

5. Requesting Information:\newline
Formal Context: meeting between employee and supervisor\newline
Example Specific Norm: Asking coworker's personal finances. Questions about a person's salary, wealth, or how much things cost are considered an invasion of privacy and very rude.\newline

Casual Context: conversation between friends\newline
Example Specific Norm: In American culture, casual requests are typically made directly and with a casual tone, often using "could you" or "would you mind" followed by the request. \newline

6. Granting a Request:\newline
Formal Context: conversation with a colleague\newline
Example Specific Norm: Agree to the request with a positive response such as, "Of course, I’d be happy to help," accompanied by a nod or smile to reinforce willingness.\newline

Casual Context: conversation between siblings\newline
Example Specific Norm: In American casual settings, granting requests are typically positive tone, often using "of course" or "no problem". \newline

7. Disagreement:\newline
Formal Context: business discussion between employees\newline
Example Specific Norm: Use respectful language such as "I see your point, but I’d like to offer another perspective," while maintaining a calm tone and open body language to show that the disagreement is friendly and constructive.\newline

Casual Context: conversation between siblings\newline
Example Specific Norm: In American culture, informal disagreements are often handled with directness and a focus on resolving the issue, rather than avoiding confrontation or resorting to indirect language. \newline

8. Agreement:\newline
Formal Context: business meeting\newline
Example Specific Norm: Use respectful language such as "I second that motion" or "I concur with that statement” for endorsement and approval. \newline

Casual Context: conversation between siblings\newline
Example Specific Norm: In casual settings with friends, agreement is usually expressed in a relaxed and informal way. Some ways of describing that can be "Yeah, totally!", "For sure!", or "Absolutely!"\newline

9. Farewells:\newline
Formal Context: business meeting\newline
Example Specific Norm: Use respectful language such as "Goodbye", "Until next time", "Farewell" or "Take care”.\newline

Casual Context: conversation between siblings\newline
Example Specific Norm: In casual settings with friends, farewell is usually expressed in a relaxed and informal way. Some ways of describing that can be "See ya!", "Bye!", or "Later!"\newline
}
&
{\tiny\ttfamily
4. 表达赞赏:\newline
正式场合： 同事之间的反馈交流\newline
示例规范： 赞美多围绕专业素养、能力或成果，如“从您身上学到了很多”、“您的专业素养令人钦佩”。会议中向上级或嘉宾致意时应起立表达敬意。\newline

非正式场合： 朋友之间的对话\newline
示例规范： 赞赏可较为轻松幽默，如“你太厉害了！”、“哇，真牛！”等。\newline

5. 请求信息:\newline
正式场合： 员工与主管之间的会议\newline
示例规范： 表达时较为委婉，例如“方便的话……”、“我想了解一下……”或“打扰一下……”，语气应礼貌客气。\newline

非正式场合： 朋友之间的对话\newline
示例规范： 常用“你知道……吗？”、“我可以问你个事吗？”等温和语气。\newline

6. 同意请求:\newline
正式场合： 与同事的交流\newline
示例规范： 常用“可以的，我会协助您完成”或“好的，我来处理”，并辅以微笑或点头。\newline

非正式场合： 兄妹或朋友之间\newline
示例规范： 常用“行啊”、“没问题”、“好说好说”等语句。\newline

7. 表达异议:\newline
正式场合： 商务讨论\newline
示例规范： 用“我理解您的意思，不过我有些不同的考虑”或“是否可以从另一个角度考虑？”表达异议，语气应平和，肢体语言开放，突出理性沟通。\newline

非正式场合： 兄妹或朋友间\newline
示例规范： 通常直接表达不同意见，如“我不这么觉得”、“你这想法不太行”。\newline

8. 表示同意:\newline
正式场合： 商务会议\newline
示例规范： 用“我同意”、“好的”、“没问题”等语言表示认同。\newline

非正式场合： 日常对话\newline
示例规范： 常用“对！”、“嗯”、“可以啊”之类词语表示认可。\newline

9. 道别:
正式场合： 商务场合结束时
示例规范： 常用“再见”、“保持联系”、“改日再聊”、“祝您一路顺风”等。

非正式场合： 朋友之间
示例规范： “拜拜”、“走啦”、“回头见”等。
}
\end{longtable}

% ===================== LONGTABLE 3 (Items 10-12) =====================
\begin{longtable}{|>{\raggedright\arraybackslash}p{0.46\textwidth}|>{\raggedright\arraybackslash}p{0.46\textwidth}|}

\hline
\textbf{US Prompt} & \textbf{Chinese Prompt} \\
\hline
\endfirsthead

\hline
\multicolumn{2}{|c|}{\textit{\textbf{Prompt Comparison (continued)}}} \\
\hline
\textbf{US Prompt} & \textbf{Chinese Prompt} \\
\hline
\endhead

\hline
\multicolumn{2}{|r|}{\textit{Continued on next page...}} \\
\hline
\endfoot

\hline
\endlastfoot

{\tiny\ttfamily
10. Rejecting a Request:\newline
Formal Context: conversation with a colleague\newline
Example Specific Norm: A formal way to reject a request is to be polite, clear, and professional. Express appreciation first, then provide a clear but polite rejection, or offer a brief explanation or suggest an alternative.\newline

Casual Context: conversation between siblings\newline
Example Specific Norm: In American casual settings, rejecting requests can be more relaxed but still polite and considerate. Some rejection might use humor if appropriate.\newline

11. <Custom category>:\newline
When the above categories do not apply, generate a new norm category. Replace the \"Custom category\" with the newly generated norm category. Do not output \"Custom category\" as the norm\_category. \newline
Example:\newline
norm\_category: Expressing criticism \newline
Casual Context: business meeting\newline
Example Specific Norm: Offer any criticism in a way that emphasizes a person’s strengths and highlights ways they could easily improve. \newline

12. No norm:\newline
When no social norm can be applied.\newline

Output Format (JSON):\newline

```json
[{
  "timestamp": {
    "start": "MM:SS",
    "end": "MM:SS"
  },
“context”: “Brief description of the setting and hierarchy between the participants”,
"norm\_category": "Category from the above list of norms",
“norm\_subject”: “Specifically to whom the norm is applied to in this context (no character name)”,
“specific\_norm”: “Specific norm in the norm category that is applicable to this context like the Example Specific Norm”,
"norm\_adherence": "adherence/violation",
"explanation": {
    "verbal\_evidence": "Description of verbal cues supporting adherence/violation.",
    "nonverbal\_evidence": "Description of nonverbal cues supporting adherence/violation."
  }
},
{
....
}]
"""
}
&
{\tiny\ttfamily
10. 拒绝请求: \newline
正式场合： 工作中回绝他人请求\newline
示例规范： 避免直接说“不”，可用“目前这个时间点不太合适”、“我需要先请示领导”等委婉方式拒绝，同时表达未来愿意协助的态度。\newline

非正式场合： 与熟人交流\newline
示例规范： “这个好像不太方便”、“我现在有点忙”、“再看看吧”。\newline

11. 自定义类别:\newline
当以上类别不适用时，**可根据情境生成一个新的社会规范类别，并将其填入 规范类别 字段**。**不能输出“自定义类别”**。\newline

示例：\newline
规范类别：提出批评\newline
非正式场合： 商务会议\newline
示例规范： 提出批评或反馈时应使用礼貌、含蓄、不伤面子的表达方式，体现对“和谐”、“人情”与“面子”的重视。\newline

12. 无规范:\newline
当场景中不存在任何适用的社会规范时使用此类别。\newline

输出格式：\newline

```json
[{
  "时间戳": {
    "开始": "MM:SS",
    "结束": "MM:SS"
  },
  "情境描述": "对场景和参与者间地位关系的简要描述",
  "规范类别": "上方规范类别之一",
  "行为主体": "此规范执行于谁（不使用具体角色名）",
  "具体规范": "在该情境中应遵循的具体社会规范",
  "规范遵循情况": "遵守 或 违反",
  "解释说明": {
    "语言证据": "支持该行为是否符合规范的语言线索",
    "非语言证据": "支持该行为是否符合规范的非语言线索"
  }
},
{
...
}]
"""
}
\end{longtable}
\twocolumn

\begin{table}[h]
\centering
\begin{tabular}{lcccccc}
\toprule
\multirow{2}{*}{\textbf{Model}} &
\multicolumn{2}{c}{\textbf{Task 1}} &
\multicolumn{4}{c}{\textbf{Task 2}} \\
\cmidrule(lr){2-3}\cmidrule(lr){4-7}
& \textbf{US} & \textbf{CN} & \textbf{US Total} & \textbf{US Correct} & \textbf{CN Total} & \textbf{CN Correct} \\
\midrule
InternVL3   & 273 & 269 & 259 & 148 & 262 &  89 \\
InternVL3.5 & 273 & 269 & 268 & 182 & 261 & 167 \\
Onevision   & 273 & 269 & 273 & 198 & 262 & 166 \\
QwenVL2     & 273 & 269 & 270 & 171 & 262 & 156 \\
QwenVL2.5   & 273 & 269 & 271 & 139 & 262 & 112 \\
Videochat   & 273 & 269 & 269 & 181 & 262 & 171 \\
llava       & 273 & 269 & 271 & 165 & 219 & 181 \\
\bottomrule
\end{tabular}
\caption{Number of evaluation samples for Task 1 and Task 2 across models.}
\label{tab:dataset_statistics}
\end{table}

\begin{table}[h]
\centering
\begin{tabular}{lcc|cc}
\toprule
\multirow{2}{*}{\textbf{Model}} &
\multicolumn{2}{c|}{\textbf{US (62 items)}} &
\multicolumn{2}{c}{\textbf{CN (39 items)}} \\
\cmidrule(lr){2-3}\cmidrule(lr){4-5}
& \textbf{Verbal} & \textbf{Nonverbal} & \textbf{Verbal} & \textbf{Nonverbal} \\
\midrule
InternVL3   & \textbf{2.79} & \textbf{2.44} & 3.46 & 2.76 \\
InternVL3.5 & 2.77  & 2.15 & 3.36 & \textbf{2.92} \\
Onevision   & 2.69 & 2.07 & \textbf{3.64} & 2.66 \\
QwenVL2     & 2.79 & 2.08 & 3.18 & 2.45 \\
QwenVL2.5   & 2.77 & 2.08 & 3.62 & 2.62 \\
Videochat   & 2.70 & 2.13 & 3.31 & 2.36 \\
llava       & 2.43 & 2.34 & 2.23 & 1.69 \\
\bottomrule
\end{tabular}
\caption{Average verbal and nonverbal scores on the intersection of items correctly predicted by all models.}
\label{tab:intersection_scores}
\end{table}
\end{document}